\documentclass[a4paper]{cas-sc}

\usepackage[authoryear]{natbib}
\usepackage{amsmath}
\usepackage{bm}
\usepackage{enumitem}
\usepackage[ruled,linesnumbered,algo2e]{algorithm2e}
\SetKw{Kwin}{$\in$}
\SetKw{Kwpara}{in parallel}
\SetKw{KwOp}{Output:} 
\SetKwRepeat{Do}{do}{while}
\newlength\myindent
\makeatletter
\newcommand{\nosemic}{\renewcommand{\@endalgocfline}{\relax}}
\newcommand{\dosemic}{\renewcommand{\@endalgocfline}{\algocf@endline}}
\makeatother
\usepackage[figuresright]{rotating}
\usepackage{amssymb}
\usepackage{mathtools}
\usepackage{multirow}
\usepackage{graphicx}
\usepackage{textcomp}
\usepackage{lipsum}
\usepackage{xcolor}
\usepackage{wrapfig}
\usepackage{subcaption}
\usepackage{caption}
\usepackage{array, booktabs, colortbl, tabularx}
\newcolumntype{P}[1]{>{\centering\arraybackslash}p{#1}}
\newcolumntype{M}[1]{>{\centering\arraybackslash}m{#1}}
\usepackage{stfloats}

\DeclareMathOperator*{\argmin}{argmin}
\DeclareMathOperator*{\minimize}{minimize}
\newcommand*{\argminl}{\argmin\limits}
\newcommand*{\minimizel}{\minimize\limits}

\def\tsc#1{\csdef{#1}{\textsc{\lowercase{#1}}\xspace}}
\tsc{DA}
\tsc{DKS}
\tsc{TB}
\tsc{ME}

\begin{document}
\let\WriteBookmarks\relax
\def\floatpagepagefraction{1}
\def\textpagefraction{.001}
\shorttitle{Initialization for large-scale airline crew pairing optimization}
\shortauthors{D. Aggarwal et~al.}

\title [mode = title]{On Initializing Airline Crew Pairing Optimization for Large-scale Complex Flight Networks}
%


\author[1]{Divyam Aggarwal}[orcid=0000-0003-0740-780X]
\ead{daggarwal@me.iitr.ac.in}

\address[1]{Department of Mechanical \& Industrial Engineering (MIED), Indian Institute of Technology Roorkee, Roorkee, Uttarakhand-247667, India}

\author[1]{Dhish Kumar Saxena}[orcid=0000-0001-7809-7744]
\cormark[1]
\ead{dhish.saxena@me.iitr.ac.in}

\author[2]{Thomas B\"ack}[orcid=0000-0001-6768-1478]
\ead{t.h.w.baeck@liacs.leidenuniv.nl}

\address[2]{Leiden Institute of Advanced Computer Science (LIACS), Leiden University, Niels Bohrweg 1, 2333 CA Leiden, the Netherlands}

\author[2]{Michael Emmerich}[orcid=0000-0002-7342-2090]
\ead{m.t.m.emmerich@liacs.leidenuniv.nl}

\cortext[cor1]{\textit{Corresponding author; Email Address: dhish.saxena@me.iitr.ac.in; Postal Address: Room No.-231, East Block, MIED, IIT Roorkee, Roorkee, Uttarakhand-247667, India; Phone: +91-8218612326}}

\begin{abstract}
Crew pairing optimization (CPO) is critically important for any airline, since its crew operating costs are second-largest, next to the fuel-cost. CPO aims at generating a set of flight sequences (\textit{crew pairings}) covering a flight-schedule, at minimum-cost, while satisfying several \textit{legality} constraints. For large-scale complex flight networks, billion-plus legal pairings (variables) are possible, rendering their \textit{offline} enumeration intractable and an exhaustive search for their minimum-cost full flight-coverage subset impractical. Even generating an \textit{initial feasible solution} (IFS: a manageable set of legal pairings covering all flights), which could be subsequently optimized is a difficult (NP-complete) problem. Though, as part of a larger project the authors have developed a crew pairing optimizer ($AirCROP$), this paper dedicatedly focuses on IFS-generation through a novel heuristic based on divide-and-cover strategy and Integer Programming.  For real-world large and complex flight network datasets (including over 3200 flights and 15 crew bases) provided by GE Aviation, the proposed heuristic shows upto a ten-fold speed improvement over another state-of-the-art approach. Unprecedentedly, this paper presents an empirical investigation of the impact of IFS-cost on the final (optimized) solution-cost, revealing that too low an IFS-cost does not necessarily imply faster convergence for $AirCROP$ or even lower cost for the optimized solution.
\end{abstract}




\maketitle

\section{Introduction} \label{intro}
\noindent For an airline, crew operating cost is the second largest component of the total operating cost (the first being the fuel cost). As a result, even marginal percentage reductions may correspond to millions of dollars annually, for a large-scale airline. With this huge potential of cost-savings, Airline Crew Scheduling (ACS) is considered as one of the most critical components of the airline scheduling process. ACS is aimed at (1) generating a low-cost set of legal/valid flight sequences covering a given flight schedule, and (2) assigning crew members on these flight sequences while fulfilling the crew requirements of these flight sequences. Some examples of key performance indicators (KPIs) used by airlines to analyze the performance of their crew scheduling processes are total deadhead\footnote{A \textit{deadhead} flight is a flight in which one crew member is actually operating the flight whereas another crew member (the deadhead crew) is travelling as a passenger.}-ing time, total number of aircraft changes performed by the crew, total overnight-rests \& sit-times, total hotel nights, etc. Airlines desire to minimize these KPIs to optimize their crew utilization. This is achieved by increasing the flying time of a crew and/or reducing their non-productive hours by reducing the time spent in sit-times, overnights, aircraft changes, etc.
\par ACS is a computationally intensive optimization problem. To maintain tractability, it is solved sequentially in two steps, namely \textit{crew pairing optimization} (CPO) and \textit{crew rostering} (also called \textit{crew assignment}). The former problem aims at generating a set of flight sequences (each called a \textit{crew pairing}) to cover a finite set of an airlines' flight schedule at minimum operating cost, while satisfying multiple \textit{legality} constraints linked to the airline-specific regulations, federations' rules, labor laws, etc. The latter problem focuses on assigning crew members to the optimal crew pairings (obtained as solution of the former problem) while satisfying underlying crew requirements. \citet{zeghal2006modeling} attempted to solve these two problems (crew pairing and crew rostering problems) in a single model but for small-scale flight networks. However, the demand based expansion of the airline operations has lead to huge-scale problems (with enormous search space and high computational requirements), rendering the single model approach intractable. This builds the rationale for solving these two problems sequentially, rather than in a single model. CPO, being the foremost problem, is the focus of this research paper and all interested readers are referred to \citet{barnhart2003airline} for an extensive review of the ACS.
\par CPO is a highly constrained NP-hard\footnote{For NP-hard (NP-complete) problems, no polynomial time algorithms on sequential computers are known up to now. However, verification of a solution might be (can be) accomplished efficiently, i.e., in polynomial time.} 
optimization problem \citep{garey2002computers}. A \textit{crew pairing} is a sequence of flights that begins and ends at the same crew base. To be operational or classified as \textit{legal}, these crew pairings have to satisfy multiple legality constraints linked to federations' safety rules, airline-specific regulations, labor laws, etc., which are discussed in the following section. For large-scale airlines (huge search space and large number of complex legality constraints), it is advisable to address CPO using two subproblems, namely \textit{crew pairing generation problem} (CPGP) and \textit{crew pairing optimization problem} (CPOP). The solution to the former subproblem (CPGP) facilitates a set of legal crew pairings required for a finite set of flights at the different stages of the airline CPO. The latter subproblem (CPOP) is solved to find a minimal-cost set of legal crew pairings, covering the given flight schedule by using advanced optimization techniques. Multiple solutions for CPGP have been proposed in literature and readers are referred to \citet{aggarwal2018large} for their extensive review. In literature, CPOP is modeled as either a \textit{set covering problem} or a \textit{set partitioning problem}. In the former problem, each flight is allowed to be covered in multiple pairings whereas in the latter problem, it is allowed to be covered only once. The adopted approaches in this research are discussed in detail in Section~\ref{model}. Legal crew pairings could be facilitated to the optimization phase in two ways: one way is to enumerate them \textit{offline}, i.e., enumerating all possible legal pairings before the optimization phase, while the other way is the \textit{online} enumeration, i.e. during the optimization phase on the requirement-basis. The former approach is advantageous for solving small-scale CPOPs ($\leq$1000 flights approximately) where enumeration and storage of all legal pairings is not computationally expensive. However, millions/billions of legal pairings are possible in large-scale CPOPs ($>$1000 flights along with multiple crew bases), rendering their offline enumeration and storage impractical. Hence, the latter approach is adopted to solve large-scale CPOPs.

\subsection{Related Work} \label{sec:relatedWork}
\noindent Depending on the problem size, mainly two types of solution approaches have been proposed in the literature: heuristic-based and mathematical programming-based solution approaches. In heuristic-based solution approaches, the most widely adapted optimization class of techniques are Genetic Algorithms (GA) which are population-based randomized-search heuristics, inspired by the theory of genetics and natural selection. Customized GAs have been widely adopted for solving combinatorial optimization problems from other domains, such as railway crew scheduling \citep{park2006crew}, resource-allocation problems \citep{deb2017population}, etc.
In the literature, the GA-based CPOP solution approaches have been proposed only for the problems for which enumeration of all legal pairings before the optimization phase is computationally tractable \citep{beasley1996genetic, levine1996application, ozdemir2001flight, kornilakis2002crew, zeren2012improved, deveci2018evolutionary, aggarwal2020realworld}. \citet{zeren2012improved} demonstrated the efficacy of GA in solving small-scale CPOPs (leading to a solution within 0.04\% gap of the global optima for a 710-flight dataset). However, \citet{zeren2016novel} demonstrated that their previous GA-based approach \citep{zeren2012improved} fails to deliver high quality solutions (around 7\% far from global optima) for large-scale CPOPs. \citet{aggarwal2020realworld} also provides a similar empirical evidence and shows that GA-based approaches (with higher-level customizations) are inefficient for solving even a small-scale yet complex flight network (839 flights). Complex flight networks are characterized by the presence of multiple hub-and-spoke subnetworks and/or multiple crew bases, leading to an explosion of possible legal pairings (millions/billions). For optimizing large-scale and complex flight networks, mathematical programming-based solution approaches have been proposed in literature.
The most widely adopted strategy in these approaches is called \textit{column generation} (CG) technique. CG is an efficient search space exploration technique which generates only the variables promising the associated cost improvements to the objective function. 
CG technique and the resulting optimization framework are not the focus of this paper. Interested readers are referred to:
\begin{itemize}
\item \citet{du1999stabilized, lubbecke2005selected, lubbecke2010column} for an extensive review of CG technique,
\item \citet{barnhart1998branch} for the review of a \textit{branch-and-price} algorithm, which is developed by integrating CG with a branch-and-bound algorithm \citep{land1960an}, and
\item \citet{anbil1991recent, anbil1992global, vance1997heuristic, anbil1998column, lavoie1988new, gustafsson1999heuristic, desaulniers2010airline, zeren2016novel, aggarwal2020aircrop} for the implementation of CG-based CPOP solution approaches.
\end{itemize}
\par To initialize these CPOP solution approaches, it is desired to generate a \textit{feasible} solution which is a manageable set of legal pairings covering all given flights. This solution is, hereby, termed as \textit{initial feasible solution} (IFS).
Given the NP-hard computational complexity of CPOP, generation of an IFS standalone is computationally challenging as it constitutes an NP-complete problem. Conventionally, graph traversal algorithms, such as depth-first search (DFS) \citep{tarjan1972depth}, are employed for traversing a connected flight graph to generate legal pairings until a feasible solution is obtained \citep{zeren2012improved}. \citet{vance1997heuristic} used artificial pairings with very high cost (one pairing to cover one flight each) for IFS generation. \citet{yan2002airline} developed a special IFS generation heuristic which involves legal pairing generation from only one-flight duties. Such IFS generation practices lead to IFSs with extremely poor cost, initializing the subsequent optimization phase from a poor initial point. This ultimately results in larger runtime of the subsequent optimization search which could be saved by generating an IFS with good cost quality. \citet{klabjan2001solving} proposed a randomized approach for IFS generation, termed as \textit{greedy randomized adaptive search procedure} (GRASP), which exploits the prior knowledge of \textit{good} flight connections. \citet{ahmadbeygi2009integer} proposed an IFS generation heuristic in which DFS algorithm is used for generation of legal pairings while allowing each flight to be covered in at most $z$ pairings (using $z=6$). \citet{aydemir2013crew} developed a \textit{knowledge based random algorithm} (KBRA) for IFS generation which exploits the knowledge of candidate flights for a preceding flight during legal pairing generation while simultaneously removing unique flights from the network as soon as they get covered. \citet{deveci2018evolutionary} generated an IFS using a knowledge-based random legal pairing generation heuristic. In that, random flight sequences are generated which are repaired using a repair-heuristic and from these repaired sequences, a high-quality IFS is searched using a GA-based optimization procedure. The majority of the above-mentioned IFS generation approaches has been validated using small-scale flight data sets (up to 800 flights) in which the size of search space does not directly affect the search-efficiency of these heuristics. Moreover, the runtime values of these IFS generation heuristics are insignificant as the search space dealt with is small in size. However, in case of large-scale CPOPs such as the ones used in this research (up to 3228 flights with 15 crew bases), the number of possible flight connections are so huge that generation of an IFS using the above-cited methods is computationally impractical, let alone their runtime values. These runtime values, if saved, could be invested in the subsequent optimization phase. In literature, two IFS generation approaches have been proposed for large-scale CPOPs, one in \citet{zeren2016novel} and the other in \citet{aggarwal2018large}. In the former approach, a large-scale CPOP for a monthly flight schedule of Turkish Airlines (flying $\sim$570 flights per day) is solved. However, a detailed procedure of its IFS-generation is not provided as a result of which it cannot be replicated for other large-scale problems. In the latter approach, the utility of the developed IFS generation heuristic is shown on a weekly flight schedule taken from a complex flight network (up to 4228 flights with 15 crew bases). It has been replicated in this research and compared with the proposed methodology. From the computational experiments, it is concluded that IFS generation approach of \citet{aggarwal2018large} is highly dependent on the characteristics of the flight datasets. This builds the rationale for the development of an IFS generation heuristic that could initialize large-scale CPOPs with complex flight networks in a computationally- and time-efficient manner.
\subsection{Contributions}
\noindent The main contributions of this paper are two fold, as follows:
\begin{itemize}
\item An iterative \textit{Integer Programming-based Divide-and-cover Heuristic} (IPDCH) is proposed, whose scalability and time-efficiency for large-scale CPOPs with complex flight networks has been demonstrated. As the name suggests, IPDCH relies on randomly decomposing the given flight data set into smaller flight subsets of pre-defined size. For each such subset, all possible legal pairings are generated and a minimum-cost subset is selected from it using the Integer Programming (IP) technique. These minimal-cost pairing subsets are combined to generate the desired IFS. 
\item An analysis of the impact of an IFS cost-quality on the runtime and final crew pairing solution quality is presented. Surprisingly, the CPP literature is quite silent on such empirical analysis which is otherwise critically important for the success of an initialization heuristic to result in a good-cost final crew pairing solution in less runtime. For this, the proposed IPDCH is run for longer times (beyond the logical termination point) to generate multiple IFSs with varying characteristics. Subsequently, $AirCROP$ is used to generate final crew pairing solutions using these IFSs individually, which are then analyzed and compared. This analysis provides an empirical evidence on how much time should be invested in the initialization phase in order to reach a near-optimal solution in a fast-manner for large and complex CPPs.
\end{itemize}
\noindent The utility of the above contributions is demonstrated on real-world, large-scale (over 3228 flights), complex flight networks (15 crew bases and multiple hub-and-spoke subnetworks), provided by GE Aviation. From the computational experiments, it is established that too low an IFS cost (using longer IPDCH-runs) does not necessarily imply lower cost of the final crew pairing solution or even its faster convergence. It is to be noted that the rationale behind this empirical study comes from unprecedented-scales and complexity of the current flight networks for which enumeration of all legal pairings is computationally impractical.
\subsection{Outline}
\noindent The outline of this paper is as follows: Section~\ref{ACP} contains an overview of airline CPO including a brief discussion on the associated terminology, pairing legality constraints, legal pairing generation approach, and CPOP modeling practices; Section~\ref{methods} includes the proposed IFS generation heuristic; Section~\ref{exp} contains the computational experiments; and lastly, Section~\ref{conc} concludes this research.

\section{Airline Crew Pairing Problem} \label{ACP}
\subsection{Terminology}
\noindent An overview of airline CPO is presented in this section. The associated terminology is as follows:
\begin{itemize}
\item The input includes a flight schedule along with the information of associated fleet and aircraft schedules (from previous scheduling steps of the airline scheduling process). 
\item The sequence of flights flown by a crew in its working day is called a \textit{crew duty} or a \textit{duty period}.
\item Any two flights in a crew duty are separated by a short time-interval, called \textit{sit-time} or \textit{connection-time}. The sit-time is required to facilitate aircraft changes for crew members between two flights, if required, or a short-rest.
\item A rest-period, longer than sit-time, is provided in between two crew duties, called as \textit{overnight-rest}, indicating the end of current duty.
\item Two short time-intervals, called as \textit{briefing-time} and \textit{debriefing-time}, are provided in starting and ending of a crew duty, respectively.
\item Total time of a crew duty, including total flying time, total sit-time, briefing and debriefing times, is called \textit{elapsed time} of the crew duty.
\item Depending on an airline's operations, some airports are selected as base of crew-operations, serving as home airports for crew members. These airports are called \textit{crew bases} or \textit{crew domiciles}.
\item A legal sequence of flights to be flown by a crew member, starting and ending at its crew base, is called a \textit{crew pairing}. An example of a crew pairing and its associated terminologies are presented in Fig.~\ref{basic}.
\begin{figure}[pos=htbp]
\centering{\includegraphics[width=0.6\columnwidth, keepaspectratio]{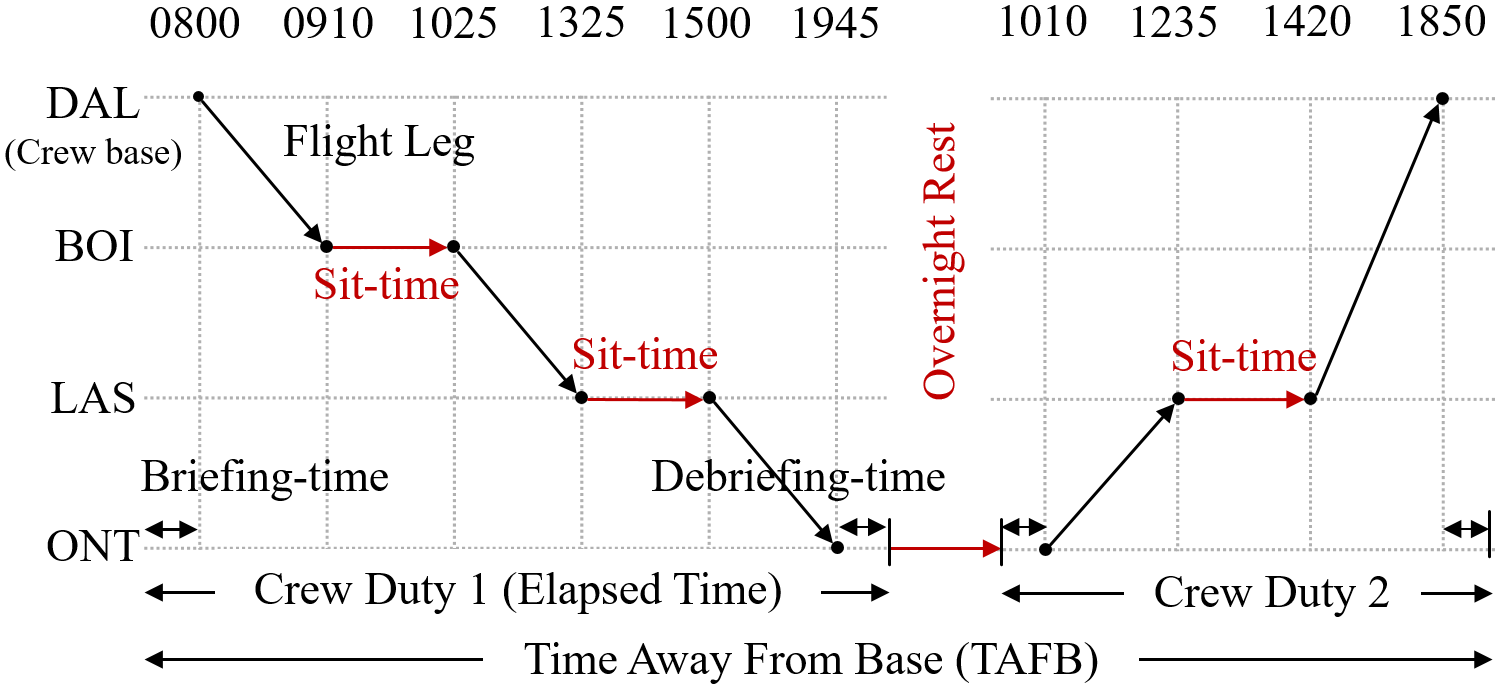}}
\centering \caption{A legal crew pairing starting from Dallas, \textit{DAL}, crew base}
\label{basic}
\end{figure}
\item The total time elapsed in a crew pairing is called \textit{time away from base (TAFB)}. It is the time-duration for which a crew is not present at its crew-base. 
\item Sometimes, flight delays and cancellations occur as an effect of unaccountable or uncertain events in real-time operations, which might result in missed flight connections for some crew members. This could also lead to a situation where crew members end up at an airport different from their crew base, and have to be transported back to their crew base in order to complete their pairing. In such cases, these crew members are transported back to their scheduled airports either using road transportation (in case of same city airports) or traveling as passengers in some other flights (in case of distant airports). The flights in which crew members travel as passengers are known as \textit{deadhead flights} or \textit{deadheads} for the traveling crew. A deadhead flight affects an airline's profits in two-folds, one by wasting the passenger seats which could have generated profits otherwise, and the other is by paying wages to the deadhead crew for a non-flying or a non-working time. Hence, airlines desire to minimize deadheads in their crew operations (ideally zero) in order to maximize their profits.
\end{itemize}

\subsection{Crew Pairing: Legality Constraints}
\noindent A crew pairing is required to satisfy several \textit{legality} constraints linked to federations' (such as FAA\footnote{FAA: Federal Aviation Administration.}, EASA\footnote{EASA: European Aviation Safety Agency.}, and other authorities) safety rules, airline-specific regulations, labor laws, etc., in order to to be classified as \textit{legal} or to become \textit{operational}. The generic form of these legality constraints is as follows:
\begin{itemize}
\item \textit{Connection-city Constraint}: In a pairing, two flights should be connected if and only if their connecting airports are same i.e. the arrival airport of the incoming flight is same as the departure airport of the outgoing flight.
\item \textit{Start-city and End-city Constraints}: A crew can start its pairing from a crew base only. Hence, the first flight of a legal pairing should start from a crew base and its last flight should end at the same crew base.
\item \textit{Sit-time \& Rest-time Constraints}: These constraints restrict the sit-time between two consecutive flights in a crew duty and the rest-time between two consecutive crew duties in a pairing. According to this constraint, only those connections are legal whose sit-time or rest-time are in between their minimum and maximum limits, respectively.
\item \textit{Duty Constraints}: These constraints restrict multiple parameters related to a crew duty. Some of the examples of these constraints are as follows:
\begin{itemize}
\item Number of duties allowed in a pairing cannot go beyond a maximum limit.
\item In a crew duty, the total number of flights, total elapsed-time and total flying-time are restricted by their maximum limits which changes according to the duty's start time.
\end{itemize}
\item \textit{Special Constraints}: Airlines formulate special constraints to further optimize their crew utilization. For example, a legal pairing is not allowed to overnight at an airport which is in the same city as that of the crew base from which it started, etc.
\end{itemize} 

\subsection{Legal Crew Pairing Generation Approach} \label{LPGP}
\noindent The number of legality constraints may vary from airline to airline as per their requirements. Hence, it is imperative to develop an efficient solution to legal crew pairing generation that could facilitate legal crew pairings to the IFS generation and optimization phases on requirement basis. 
In \citet{chu1997solving}, it is recorded that generation of legal pairings throughout the airline CPO consumes 25-50\% of its total runtime, making it imperative to develop a computationally- and time-efficient solution to CPGP. A plausible approach could be to exploit the priority order of legality constraints while generating the legal pairings. This might eliminate the need for checking lesser priority constraints if higher ones are already illegal, hence, saving the runtime to an extent. Moreover, decomposition of CPGP into smaller subproblems and adoption of parallel processing techniques to solve them will speed-up the overall process. Some of the existing parallel architectures have been discussed in \citet{aggarwal2018large}. However, such parallel architectures require expensive computational resources. Furthermore, the efforts required in implementation of such complex concepts of parallelization may hamper the development and testing of new ideas, becoming a barrier-to-entry for new researchers. Towards this, \citet{aggarwal2018large} proposed a simpler yet parallel legal crew pairing generation approach, enabling new researchers for developing and testing new ideas in a simpler and faster pairing generation environment. In large-scale CPOPs, multiple crew bases are present along with complex pairing legality constraints which may vary with-respect-to the crew bases from which a pairing starts. Exploiting this, authors decomposed the legal pairing generation process into independent subprocesses with-respect-to each crew base. These subprocesses are executed, in parallel, on idle cores of a single processing unit. The above legal pairing generation approach has been adopted in this research for facilitating legal crew pairings to the proposed IFS generation heuristic in a computationally- and time-efficient manner. 

\subsection{Crew Pairing Optimization Problem Formulation} \label{model}
\noindent In literature, the crew pairing optimization problem has been modeled as either a set partitioning problem or a set covering problem. These problem formulations fulfill the basic modeling requirements of CPOP. As mentioned in the introduction, in a set partitioning model, each flight is allowed to be covered only once (no deadhead flights are allowed), whereas, in a set covering model, flights are allowed to be covered in more than one pairing (over-coverage represents deadhead flights). Although a set partitioning problem formulation might result in an improved solution, it might also lead to an infeasible solution in problems where an ideally-partitioned solution, i.e., a set of legal pairings covering all flights only once (without deadheads), does not exist. Moreover, the set covering model might result in faster convergence in case of large-scale CPOPs \citep{gustafsson1999heuristic}, making it a more appropriate modeling choice. In the context of CPOP, each row of the constraint matrix represents a flight and each column represents a legal pairing. In this research, a set covering problem formulation is used to model crew pairing optimization subproblems in the proposed IFS generation heuristic (IPDCH) as presented in the next section. For a given set of legal pairings, $\mathcal{P}$, covering a set of flights, $\mathcal{F}$, the set covering problem formulation (labeled CPOP) is as follows.
\begin{flalign}
(\text{CPOP}) \qquad &\minimizel_{\mathbf{x}}~\text{f}(\mathbf{x}) = \left(\sum_{j=1}^{P} c_j x_j +\left(\sum_{i=1}^{F}\left(\sum_{j=1}^{P}  a_{ij} x_{j} - 1 \right) \right) \times P_{Dhd}\right), &\label{eq:obj}\\
&\text{subject to} \quad \sum_{j=1}^{P} a_{ij} x_{j} \geq 1,~~~~\forall i \in \{1,2,...,F\} \label{eq:coverage}\\
&\qquad \qquad \quad x_j \in \{0, 1\},~~~~~~\forall j \in \{1,2,...,P\} \label{eq:integrality}
\end{flalign}
\begin{flalign}
\qquad \qquad \quad \text{where} \qquad~~P &:~\text{size of }\mathcal{P}\text{, i.e., } |\mathcal{P}|,\nonumber \\
\qquad \qquad \quad F &:~\text{size of }\mathcal{F}\text{, i.e., }|\mathcal{F}|,\nonumber \\
\qquad \qquad \quad c_j &:~\text{cost of a pairing }p_j,\nonumber \\
\qquad \qquad \quad P_{Dhd} &:~\text{airline-defined parameter which penalizes the number of deadheads in the solution},\nonumber \\
\qquad \qquad \quad a_{ij} &:~ 
 \begin{cases}
 1, & \text{if flight}~f_i~\text{is covered in pairing}~p_j\\
 0, & \text{otherwise}
 \end{cases}, \nonumber \\
\qquad \qquad \quad x_j &:~
 \begin{cases}
 1, & \text{if pairing}~p_j~\text{is selected in the corresponding solution}\\
 0, & \text{otherwise}
 \end{cases}, \nonumber \\
\qquad \qquad \quad \mathbf{x} &:~ \left[x_1~x_2~x_3~...~x_P\right]^{\mathsf{T}} \nonumber
\end{flalign}

Eq.~\eqref{eq:obj} represents the objective function which is to minimize the total cost of the selected pairings. Eq.~\eqref{eq:coverage} represents a set of $F$ flight-coverage constraints which ensures that all $F$ flights are covered by at least one legal pairing. Eq.~\eqref{eq:integrality} represents a set of $P$ side-constraints which ensures the binary nature of all $P$ decision variables.

\section{Proposed Methodology} \label{propMeth}
\subsection{Initial Feasible Solution and Its Associated Characteristics} \label{methods}
\noindent A \textit{feasible} crew pairing solution is a set of legal pairings covering all given flights, satisfying the set of $F$ flight-coverage constraints (Eq.~\eqref{eq:coverage}). As discussed in Section~\ref{intro}, for large-scale CPOPs, the optimization phase is required to be initialized using a feasible, yet manageable, set of legal pairings. This pairing set is termed as an \textit{initial feasible solution} (IFS). For an IFS, two important characteristics are identified whose interplay is critical in driving the subsequent optimization search towards a global optimum. These are as follows:
\begin{itemize}
\item \textit{Cost}: It is measured as the cost of a linear programming (LP)-solution obtained from the generated IFS. This LP-solution is obtained by solving CPOP with a continuously-relaxed form of the decision variables, i.e., $x_j \in [0,1]\ \ \forall j \in [1,2,...,P]$, instead of Eq.~\eqref{eq:integrality}. The LP-cost of the generated IFS could be regulated by timing the termination of the proposed IFS generation heuristic. In that, as IPDCH is kept running, more pairings are generated and added to the desired IFS which might either lower its LP-cost or it remains constant.
\item \textit{Degrees of search-freedom:} In large-scale optimization problems, the interplay between exploration and exploitation by an optimization algorithm determines its search-efficiency, as presented in fig.~\ref{fig:tradeoff}.
\begin{figure}[pos=htbp]
\centering
\includegraphics[width=0.40\columnwidth, keepaspectratio]{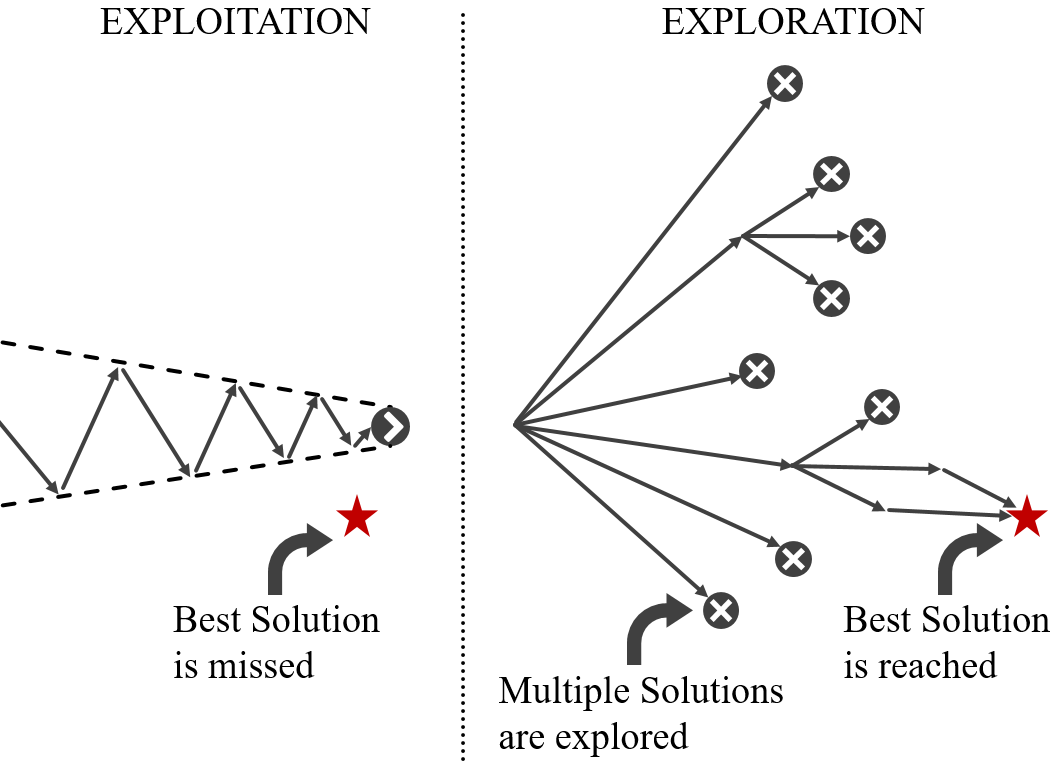}
\centering \caption{Exploratory versus exploitative optimization search \citep{des2019two}.}
\label{fig:tradeoff}
\end{figure}
In order to prevent local optimality, exploration is desired upfront during the initial search, and exploitation is desired subsequently as the search narrows down towards the global optimum. Since an IFS is used to initialize the optimization phase, it is desirable to generate an IFS which has flexibility for an exploratory-search rather than an exploitative-search. This flexibility of an IFS, which determines the search-direction of the subsequent optimization phase, is hereby termed as \textit{degrees of search-freedom} (DOSF). This characteristic is also regulated by timing the termination of the proposed heuristic but in a complimentary manner as that of its LP-cost. In longer IPDCH-runs, more number of pairings are added which increases the rigidity of the resulting IFS towards an exploratory-search, contrary to its LP-cost. Hence, IFSs resulting from shorter IPDCH-runs will have higher DOSF values than those from the longer IPDCH-runs.
\end{itemize}
\par The above-mentioned IFS characteristics are conflicting in nature as evident from their relationship with the termination choices of an IPDCH-run. The majority of IFS generation methods, proposed in literature, have been used for initializing small-scale CPOPs in which the interplay of these IFS characteristics remains insignificant in finding the final crew pairing solution. However, it is shown in this research that the success of subsequent optimization phase for a large-scale CPOP is highly dependent on the interplay of these IFS characteristics. As mentioned in Section~\ref{sec:relatedWork}, there are two instances in the literature which have addressed the initialization of large-scale CPOPs, one is \citet{zeren2016novel} and the other is \citet{aggarwal2018large}. In the former instance, a special initialization heuristic is used in which a DFS algorithm is used to traverse the duty-based network for generating legal pairings and a flight is not covered again if it has been covered once. A detailed procedure or a pseudo-code of this initialization heuristic is not provided, leaving its adaptability for other large-scale problems questionable. In the latter approach, an \textit{Enhanced-DFS} heuristic is proposed which is a modification of the DFS algorithm and attempts to cover unique flights by varying its backtracking step-length from the child flight nodes to the parent flight nodes. The above Enhanced-DFS is replicated in this research and compared with the proposed IPDCH. Moreover, both of the above-cited approaches did not discuss about the sensitivity of an IFS characteristics on the CPO which has been found critically important for an efficient subsequent optimization-search.

\subsection{Integer Programming based Divide-and-cover Heuristic (IPDCH)} 
\noindent The motivation behind the development of this heuristic is to generate an IFS in not only a time-efficient manner but also in a cost-effective manner for all types of CPOPs. Towards this, a divide-and-cover strategy is adopted to develop an efficient heuristic, termed as \textit{Integer Programming-based Divide-and-cover Heuristic} (IPDCH). The pseudo code of IPDCH is given in Algorihm~\ref{algo:method1} and its working is explained as follows.
\label{ipdch}
	\begin{algorithm2e}[htbp]
		\DontPrintSemicolon
		\footnotesize
		\SetKwComment{Comment}{$\triangleright$\ }{}
		\KwIn {$\mathcal{F}$; $K$; and $\texttt{Pairing\_Gen()}$\;}
		\KwOp {$\mathcal{IFS}$\;}
		\textbf{Procedure:}\;
		$\mathcal{IFS} \leftarrow \phi$, $\mathcal{\tilde{F}} \leftarrow \mathcal{F}$, $flag \leftarrow False$\;
		\While{termination criterion is not met}{
			$\mathcal{F}_{K}, \mathcal{\grave{F}}_{K'}, \mathcal{P}_{K'}, \mathbf{x}^*, \mathcal{P}_{K'}^{*} \leftarrow \phi$\;
			\uIf{$|\mathcal{F}| > K$}{
				$\mathcal{F}_K \leftarrow $ Select $K$ random flights from $\mathcal{F}$\;
				$\mathcal{F} \leftarrow \mathcal{F} \setminus \mathcal{F}_{K}$\;
			}
			\Else{
				$\mathcal{F}_K \leftarrow \mathcal{F}$\;
				$flag \leftarrow True$\;
			}
			$\mathcal{P}_{K'} \leftarrow \ $\texttt{Pairing\_Gen($\mathcal{F}_K$)}\;
			$\mathcal{\grave{F}}_{K'} \leftarrow $ Flights covered in $\mathcal{P}_{K'}$ \Comment*[f]{$K' \leq K$}\;
			Formulate CPOP using $\mathcal{\grave{F}}_{K'}$ and $\mathcal{P}_{K'}$\;
			$\mathbf{x^{*}} \gets \argminl_{\mathbf{x}}~ \text{f}(\mathbf{x})$\Comment*[f]{Solved using an IP-solver}\;
			$\mathcal{P}_{K'}^{*} \leftarrow \{p_j~|~(p_j \in \mathcal{P}_{K'}) \land (x^*_{j'}=1) \land (j=j') \}$\;
			$\mathcal{IFS} \leftarrow \mathcal{IFS} \cup \mathcal{P}_{K'}^{*}$\;
			\uIf{$flag=False$}{
				$\mathcal{F} \leftarrow \mathcal{F} \cup \{\mathcal{F}_K \setminus \mathcal{\grave{F}}_{K'}\}$\;
			}
			\Else{
				$\mathcal{F} \gets \mathcal{\tilde{F}}$\;
				$flag \leftarrow False$\;
		}}
		\textbf{return} $\mathcal{IFS}$\;
		\caption{Pseudo-code of the proposed IPDCH}
		\label{algo:method1}
	\end{algorithm2e}
\par As shown in Algorihm~\ref{algo:method1}, the input to the IPDCH includes a flight schedule, $\mathcal{F}$; a pre-defined decomposition parameter, $K$; and an efficient legal crew pairing generation sub-routine, $\texttt{Pairing\_Gen()}$. This sub-routine receives a finite set of flights as input and returns an enumerated set of all possible legal pairings for the input flight set. It has been adopted from \citet{aggarwal2018large}, as discussed in Section~\ref{LPGP}. The output of the IPDCH is the desired IFS, denoted by $\mathcal{IFS}$. IPDCH is an iterative heuristic and the working of each of its iterations is explained in lines 4 to 24. In the first IPDCH-iteration, the input flight schedule $\mathcal{F}$ is randomly decomposed into a smaller flight subset, denoted by $\mathcal{F}_K$, without replacement (lines 6 to 12). The size of $\mathcal{F}_K$ is controlled using the decomposition parameter $K$. Subsequently, in line 13, the flights in $\mathcal{F}_K$ are fed as input to $\texttt{Pairing\_Gen()}$ sub-routine to generate the set of all possible legal pairings, denoted by $\mathcal{P}_{K'}$. It is to be noted the set of flights covered in $\mathcal{P}_{K'}$, denoted by $\mathcal{\grave{F}}_{K'}$, might be a $\subseteq \mathcal{F}_K$, containing only $K'$ flights with $K' \leq K$ (line 14). The rationale behind this is the fact that there might be missed flight connections for some of the flights in $\mathcal{F}_K$ due to the random selection. In line 19, the remaining (uncovered) flights in the set $\mathcal{F}_{K} \setminus \mathcal{\grave{F}}_{K'}$ are added back to $\mathcal{F}$ to ensure fair-chance of their selection in subsequent IPDCH-iterations. In line 15, a CPOP is formulated using the generated legal pairing set $\mathcal{P}_{K'}$ and its flight set $\mathcal{\grave{F}}_{K'}$ as search space (CPOP model is discussed in Section~\ref{model}). The resulting model is optimized using an Integer Programming (IP) technique, such as the branch-and-bound algorithm, and an optimal solution vector, denoted by $\mathbf{x}^*$, is obtained (line 16). Subsequently, in line 17, the optimal solution $\mathbf{x^{*}}$ is used to create the IP-optimal pairing set, denoted by $\mathcal{P}_{K'}^{*}$, by selecting only those pairings ($p_j$) from $\mathcal{P}_{K'}$ for which the corresponding $x_j^*$ is 1. The optimal pairing set  $\mathcal{P}_{K'}^{*}$ is then added to $\mathcal{IFS}$ (line 17). The subset $\mathcal{F}_K$ for the subsequent IPDCH-iteration is formed from the remaining flights, given in the set $\mathcal{F}\setminus \mathcal{\grave{F}}_{K'}$. It is to be noted that for flight selection without replacement, a global copy of $\mathcal{F}$, denoted by $\mathcal{\tilde{F}}$, is maintained to replace flights in $\mathcal{F}$ whenever its size becomes $\leq K$ (lines 3, 9-12 \& 20-23). This process is repeated until the termination criterion is met, and the $\mathcal{IFS}$ obtained after the termination of IPDCH becomes the desired IFS. An IPDCH-run could be terminated as soon as all flights in $\mathcal{F}$ are covered in at least one pairing of the $\mathcal{IFS}$, and this point is termed as the \textit{feasibility point}. Moreover, an IPDCH-run could be allowed to keep running beyond the feasibility point. More pairings would be added with each extra IPDCH-iteration which, in turn, will improve the LP-cost and decrease the DOSF of the $\mathcal{IFS}$. However, it is imperative to find the perfect balance between the characteristics of an IFS and the time spent in its generation. For this, an empirical study is presented in Section~\ref{sensitivity}.
\par The proposed IPDCH is advantageous over existing methods in two ways. First, it is utilizes a divide-and-cover strategy which drastically improves its search-efficiency. As a result, it drastically decreases the IFS generation runtime for not only large-scale CPOPs but for complex flight networks too which was the major limitation of the traditional IFS generation methods. Second, beyond covering the given flight schedule in a decomposed manner, IPDCH optimizes the decomposed flight subsets using IP in each of its iteration. As a result, only optimal pairings out of all possible legal pairings are added to the desired IFS in each iteration, bringing in the associated cost benefits.

\section{Computational Experiments} \label{exp}
\subsection{Experimental Settings} \label{sec:settings}
In this section, computational experiments are presented in order to demonstrate the utility of the proposed IPDCH and to discuss the impact of IFS-characteristics on the final crew pairing solution. In this research, the proposed/ replicated algorithms are implemented using \textit{Python 3.6} scripting language. All computations are performed on a HP Z640 workstation powered by $2\ \times$ (Intel$^\circledR$ Xeon$^\circledR$ E5-2630v3 Processors having 16 cores at 2.40GHz), and with 64GB RAM. For solving the optimization subproblems in each IPDCH-iteration, an IP-solver from Gurobi Optimizer Suit v8.1 \citep{gurobi} is used which is available for research purposes via academic license. The real-world airline test-cases, used in this study, are introduced in the following. Afterwards, the experimental results of IFS generation using the proposed IPDCH and the Enhanced-DFS heuristic are presented and compared. Lastly, the sensitivity of IFS-characteristics on the final crew pairing solution is analyzed. For this sensitivity analysis, a Column Generation based Airline Crew Pairing Optimizer ($AirCROP$) is used which has been developed by the authors and has been validated by GE Aviation \citep{aggarwal2020aircrop}. This optimization framework is used as a black-box to which the generated IFSs (using different settings of the proposed IPDCH) are fed as input and the corresponding final crew pairing solutions are received as output.

\subsection{Airline Test Cases} \label{data}
In this work, large-scale airline test cases are used which have been provided by GE Aviation. Details of these test cases are given in Table~\ref{testCase}.
\begin{table*}[pos=htb]
\scriptsize
\begin{center}
\begin{tabular}{ccccc}
\toprule
\textbf{Test} & \textbf{\# of} & \textbf{\# of Crew} & \textbf{Total Legal} & \textbf{Average \# of Duties}\\
\textbf{Cases} & \textbf{Flights} & \textbf{Base} & \textbf{Duties} & \textbf{Per Crew Base}\\
\midrule
TC1 & 2,453 & 06 & 110,348 & 18,392\\
TC2 & 3,202 & 15 & 454,205 & 30,281\\
TC3 & 3,228 & 15 & 464,092 & 30,940\\
\bottomrule
\end{tabular}
\caption{Real-world airline test cases}
\label{testCase}
\end{center}
\end{table*}
It is to be noted that TC2 \& TC3 have approximately the same number of flights. However, TC2 contains some flight legs with rare legal flight connections, making it difficult for the search algorithm to find an IFS and/or near-optimal final crew pairing solutions. These test cases have been used in this research to establish/validate the generality of the proposed IPDCH. The majority of large-scale flight data sets used in \citet{parmentier2019aircraft} consist of flights between 600-1000, excluding two flight data sets which contain 1766 \& 3398 flights respectively. Hence, the test cases used in this research could also be classified as large-scale problems as they not only contain a larger number of flights, but they are also derived from complex flight networks (multiple hub-and-spoke subnetworks and multiple crew bases). In each of these test cases, a set of pairing legality constraints and costing rules are provided along with an input flight schedule. These input flight schedules contain flight information/attributes such as its departure/arrival airports, departure/arrival time-stamps\footnote{A \textit{time-stamp} gives information about the time and day of an event.}, aircraft's fleet, and block-time\footnote{\textit{Block-time} is the total flying-time of a flight and is measured as the time taken by the flight's aircraft from departure gate of its departure airport to the arrival gate of its arrival airport.}. The pairing legality constraints have been discussed in Section~\ref{ACP}. However, the exact parameters and some airline-specific constraints cannot be revealed due to confidentiality agreements with the industrial sponsor. A non-linear set of costing rules is provided for computation of the cost of a legal pairing. According to that, the pairing cost is made up of two components namely, a \textit{flying cost} component and a \textit{variable} cost (or \textit{non-flying} cost). The flying cost is the total cost of flying all given flights (excluding deadhead flights). The variable cost is the total cost incurred by an airline during non-flying hours of a crew in a pairing, which is further divided into following components:
\begin{itemize}
\item \textit{Hard cost}: It is made up of \textit{excess pay} (or \textit{credit pay}), and the hotel \& meal cost. Excess pay is the cost associated with non-productive hours of a crew in a pairing, and is calculated as the cost of all minimum guaranteed hours that are not met by the actual flying hours. These minimum crew guaranteed hours depend upon various attributes of a pairing such as its duty elapsed-time, TAFB, deadhead flights, etc. Airlines desire to minimize this cost (ideally to zero) in order to maximize their crew utilization. Hotel \& meal cost is the cost of accommodation during overnight-rest periods and meal expenses for the entire pairing.
\item \textit{Soft cost}: It is the penalty cost added for every aircraft change while making a flight connection in a pairing, and for every deadhead flight used in a pairing.
\end{itemize}
During discussions with domain experts from GE Aviation, it is learnt that airlines primarily use hard cost as the sole KPI for evaluating the performance of their CPO process. 
\subsection{Results of IFS Generation} \label{ifsresults}
\subsubsection{IFS generation using the proposed IPDCH}
It is imperative to discuss the experimental settings of the proposed IPDCH first. The proposed IPDCH involves a decomposition parameter, $K$, whose function is to decompose the large-scale input flight schedule into smaller flight subsets. The primary objective of the proposed IPDCH is to generate an IFS for large-scale and complex CPOPs in a time-efficient manner. Hence, the best-setting of $K$ should be selected in a way to balance the trade-off between:
\begin{itemize}
\item the size of the decomposed flight subset which, in turn, affects the presence of legal flight connections for the selected flights, and
\item the runtime required in pairing generation as well as in solving the resulting IP for the decomposed flight subset, i.e., the runtime of an IPDCH-iteration.
\end{itemize}
Higher values of $K$ would promote legal flight connections for a larger number of flights in the selected flight subset (larger $\mathcal{\grave{F}}_{K'}$). This, in turn, will lead to generation of more legal pairings (larger $\mathcal{P}_{K'}$) . However, this will also lead to higher runtime values required for generation of $\mathcal{P}_{K'}$ and the subsequent optimization using IP, increasing the overall runtime of the IPDCH-iterations. On the contrary, lower values of $K$ would promote smaller $\mathcal{P}_{K'}$, smaller runtime for its generation, and optimization using IP. However, it may drastically reduce the probability of finding legal flight connections while generation of legal pairings, i.e., smaller $\mathcal{\grave{F}}_{K'}$. Hence, it is imperative to tune the setting of $K$ for the given computational resources (Section~\ref{sec:settings}) in order to leverage-in the maximum gain in the search-efficiency of the proposed IPDCH. For this, multiple IPDCH-runs with $K$ varying from $400$ to $1200$ flights and a $\Delta K=100$ flights are performed. The rationale behind selection of such settings of $K$ is as follows:
\begin{table*}[pos=htbp]
\scriptsize 
\begin{center}
\resizebox{\textwidth}{!}{\begin{tabular}{M{2mm}M{2mm}M{15mm}M{22mm}M{7mm}M{7mm}M{7mm}M{7mm}M{7mm}M{7mm}M{7mm}M{7mm}M{7mm}M{7mm}}
\toprule
\textbf{Test} & \multirow{2.5}{*}{\textbf{K}} & \multirow{2.5}{*}{\textbf{Measures}} & \multicolumn{11}{c}{\textbf{Pseudo-random number seeds (including mean $\pm$ standard deviation values)}} \\ 
\cmidrule(l){4-14}
\textbf{Case} & & & $\pmb{\overline x \pm \sigma}$ & \textbf{S1} & \textbf{S2} & \textbf{S3} & \textbf{S4} & \textbf{S5} & \textbf{S6} & \textbf{S7} & \textbf{S8} & \textbf{S9} & \textbf{S10} \\ 
\toprule
\multirow{23}{*}{\textbf{TC1}} & \multirow{2}{*}{\textbf{400}} & \textbf{Uncov. Flts.} & 0.0 $\pm$ 0.0 & 0 & 0 & \cellcolor{red!20}0 & 0 & 0 & 0 & 0 & \cellcolor{green!25}0 & 0 & 0 \\ 
 &  & \textbf{Runtime (s)} & 312.64 $\pm$ 165.55 & 334.0 & 217.19 & \cellcolor{red!20}706.56 & 316.36 & 188.42 & 508.64 & 295.34 & \cellcolor{green!20}123.95 & 246.33 & 189.61 \\ 
\cmidrule(l){2-14}  
 & \multirow{2}{*}{\textbf{500}} & \textbf{Uncov. Flts.} & 0.0 $\pm$ 0.0 & 0 & 0 & \cellcolor{red!20}0 & 0 & 0 & 0 & 0 & 0 & 0 &\cellcolor{green!25} 0 \\ 
 &  & \textbf{Runtime (s)} & 122.13 $\pm$ 54.99 & 152.06 & 102.75 & \cellcolor{red!20}257.25 & 173.58 & 90.81 & 79.81 & 91.82 & 85.74 & 120.53 & \cellcolor{green!25}66.98 \\ 
\cmidrule(l){2-14}  
 & \multirow{2}{*}{\textbf{600}} & \textbf{Uncov. Flts.} & 0.0 $\pm$ 0.0 & 0 & \cellcolor{green!25}\textbf{0} & 0 & 0 & 0 & 0 & 0 & 0 & \cellcolor{red!20}0 & 0 \\ 
 &  & \textbf{Runtime (s)} & 99.38 $\pm$ 55.75 & 63.82 & \cellcolor{green!25}\textbf{51.42} & 62.85 & 57.65 & 76.6 & 160.93 & 123.44 & 110.12 & \cellcolor{red!20}231.63 & 55.32 \\ 
\cmidrule(l){2-14}  
 & \multirow{2}{*}{\textbf{700}} & \textbf{Uncov. Flts.} & 0.0 $\pm$ 0.0 & 0 & 0 & \cellcolor{green!25}\textbf{0} & 0 & 0 & \cellcolor{red!20}0 & 0 & 0 & 0 & 0 \\ 
 &  & \textbf{Runtime (s)} & 119.83 $\pm$ 71.62 & 99.72 & 87.28 & \cellcolor{green!25}\textbf{52.98} & 85.83 & 69.79 & \cellcolor{red!20}307.71 & 78.03 & 143.97 & 94.57 & 178.4 \\ 
\cmidrule(l){2-14}  
 & \multirow{2}{*}{\textbf{800}} & \textbf{Uncov. Flts.} & 0.0 $\pm$ 0.0 & 0 & \cellcolor{green!25}0 & 0 & \cellcolor{red!20}0 & 0 & 0 & 0 & 0 & 0 & 0 \\ 
 &  & \textbf{Runtime (s)} & 221.57 $\pm$ 112.79 & 168.82 & \cellcolor{green!25}73.41 & 235.75 & \cellcolor{red!20}388.11 & 75.44 & 132.03 & 340.85 & 293.37 & 145.13 & 362.76 \\ 
\cmidrule(l){2-14}  
 & \multirow{2}{*}{\textbf{900}} & \textbf{Uncov. Flts.} & 0.0 $\pm$ 0.0 & 0 & 0 & \cellcolor{green!25}0 & 0 & 0 & \cellcolor{red!20}0 & 0 & 0 & 0 & 0 \\ 
 &  & \textbf{Runtime (s)} & 166.6 $\pm$ 56.10 & 168.26 & 127.43 & \cellcolor{green!25}114.38 & 155.65 & 208.82 & \cellcolor{red!20}316.78 & 162.6 & 136.77 & 128.7 & 146.63 \\ 
\cmidrule(l){2-14}  
 & \multirow{2}{*}{\textbf{1000}} & \textbf{Uncov. Flts.} & 0.0 $\pm$ 0.0 & 0 & 0 & 0 & \cellcolor{green!25}0 & 0 & 0 & 0 &\cellcolor{red!20} 0 & 0 & 0 \\ 
 &  & \textbf{Runtime (s)} & 707.15 $\pm$ 500.84 & 363.59 & 309.88 & 934.55 &\cellcolor{green!25} 227.23 & 400.08 & 1218.4 & 312.92 & \cellcolor{red!20}1626.62 & 310.4 & 1367.84 \\ 
\cmidrule(l){2-14}  
 & \multirow{2}{*}{\textbf{1100}} & \textbf{Uncov. Flts.} & 0.0 $\pm$ 0.0 & 0 & 0 & \cellcolor{green!25}0 & 0 & 0 & 0 & \cellcolor{red!20}0 & 0 & 0 & 0 \\ 
 &  & \textbf{Runtime (s)} & 1607.93 $\pm$ 994.59 & 850.7 & 536.09 & \cellcolor{green!25}531.43 & 1753.08 & 1947.7 & 2350.13 & \cellcolor{red!20}3656.11 & 2657.52 & 997.42 & 799.16 \\ 
\cmidrule(l){2-14}  
 & \multirow{2}{*}{\textbf{1200}} & \textbf{Uncov. Flts.} & 0.0 $\pm$ 0.0 & 0 & 0 & 0 & 0 & 0 & 0 & \cellcolor{green!25}0 & 0 & \cellcolor{red!20}0 & 0 \\ 
 &  & \textbf{Runtime (s)} & 3560.13 $\pm$ 1124.55 & 3584.73 & 3486.83 & 1977.61 & 4424.67 & 3668.92 & 3665.82 & \cellcolor{green!25}1785.83 & 2673.65 & \cellcolor{red!20}5533.75 & 4799.51 \\ 
\toprule
\multirow{23}{*}{\textbf{TC2}} & \multirow{2}{*}{\textbf{400}} & \textbf{Uncov. Flts.} & 241.10 $\pm$ 168.56 & 407 & 402 & 58 & 408 & 415 & 122 & 11 & 402 & 126 & 60 \\ 
&  & \textbf{Runtime (s)} & 7200.37 $\pm$ 0.22 & 7200.38 & 7200.17 & 7200.31 & 7200.55 & 7200.01 & 7200.72 & 7200.34 & 7200.66 & 7200.38 & 7200.15 \\ 
\cmidrule(l){2-14} 
& \multirow{2}{*}{\textbf{500}} & \textbf{Uncov. Flts.} & 0.0 $\pm$ 0.0 & \cellcolor{green!25}0 & 0 & 0 & \cellcolor{red!20}0 & 0 & 0 & 0 & 0 & 0 & 0 \\ 
&  & \textbf{Runtime (s)} & 790.06 $\pm$ 613.49 & \cellcolor{green!25}239.39 & 359.88 & 1245.15 & \cellcolor{red!20}2397.71 & 344.49 & 376.76 & 791.92 & 461.25 & 861.62 & 822.47 \\ 
\cmidrule(l){2-14}  
& \multirow{2}{*}{\textbf{600}} & \textbf{Uncov. Flts.} & 0.0 $\pm$ 0.0 & \cellcolor{green!25}0 & 0 & 0 & 0 & \cellcolor{red!20}0 & 0 & 0 & 0 & 0 & 0 \\ 
&  & \textbf{Runtime (s)} & 500.14 $\pm$ 286.00 & \cellcolor{green!25}120.8 & 705.14 & 263.47 & 483.1 & \cellcolor{red!20}926.95 & 131.55 & 690.34 & 185.25 & 750.57 & 744.28 \\ 
\cmidrule(l){2-14} 
& \multirow{2}{*}{\textbf{700}} & \textbf{Uncov. Flts.} & 0.0 $\pm$ 0.0 & 0 & \cellcolor{red!20}0 & 0 & 0 & 0 & 0 & \cellcolor{green!25}\textbf{0} & 0 & 0 & 0 \\ 
&  & \textbf{Runtime (s)} & 636.74 $\pm$ 750.15 & 132.92 & \cellcolor{red!20}2530.57 & 282.58 & 1467.64 & 128.38 & 212.54 & \cellcolor{green!25}\textbf{109.6} & 698.65 & 667.63 & 136.89 \\ 
\cmidrule(l){2-14} 
& \multirow{2}{*}{\textbf{800}} & \textbf{Uncov. Flts.} & 0.0 $\pm$ 0.0 & 0 & 0 & 0 & 0 & 0 & \cellcolor{green!25}\textbf{0} & 0 & 0 & 0 & \cellcolor{red!20}0 \\ 
&  & \textbf{Runtime (s)} & 684.60 $\pm$ 812.28 & 254.0 & 183.39 & 726.99 & 567.8 & 474.87 & \cellcolor{green!25}\textbf{106.75} & 294.26 & 789.58 & 412.15 & \cellcolor{red!20}3036.25 \\ 
\cmidrule(l){2-14} 
& \multirow{2}{*}{\textbf{900}} & \textbf{Uncov. Flts.} & 0.0 $\pm$ 0.0 & 0 & 0 & \cellcolor{red!20}0 & 0 & 0 & 0 & 0 & 0 & \cellcolor{green!25}0 & 0 \\ 
&  & \textbf{Runtime (s)} & 985.23 $\pm$ 785.11 & 814.26 & 492.78 & \cellcolor{red!20}2667.15 & 359.92 & 920.17 & 295.23 & 599.08 & 2131.75 & \cellcolor{green!25}216.4 & 1355.59 \\ 
\cmidrule(l){2-14} 
& \multirow{2}{*}{\textbf{1000}} & \textbf{Uncov. Flts.} & 0.2 $\pm$ 0.6 & 2 & 0 & 0 & 0 & 0 & 0 & 0 & 0 & 0 & 0 \\ 
&  & \textbf{Runtime (s)} & 2611.39 $\pm$ 1837.04 & 7219.48 & 4187.71 & 2756.86 & 419.3 & 2113.89 & 842.02 & 2405.43 & 2564.58 & 1317.88 & 2286.76 \\ 
\cmidrule(l){2-14} 
& \multirow{2}{*}{\textbf{1100}} & \textbf{Uncov. Flts.} & 0.6 $\pm$ 0.92 & 0 & 0 & 0 & 2 & 0 & 2 & 0 & 2 & 0 & 0 \\ 
& & \textbf{Runtime (s)} & 5453.02 $\pm$ 1879.96 & 5466.36 & 3064.13 & 7166.73 & 7217.62 & 4517.77 & 7245.06 & 2097.01 & 7219.2 & 3679.45 & 6856.88 \\ 
\cmidrule(l){2-14} 
& \multirow{2}{*}{\textbf{1200}} & \textbf{Uncov. Flts.} & 2.8 $\pm$ 1.6 & 0 & 5 & 3 & 3 & 2 & 2 & 3 & 2 & 6 & 2 \\ 
&  & \textbf{Runtime (s)} & 8179.58 $\pm$ 1036.51 & 6862.2 & 10503.73 & 8095.02 & 8197.77 & 7490.89 & 7711.8 & 7854.07 & 8320.46 & 9541.72 & 7218.14 \\ 
\toprule 
\multirow{23}{*}{\textbf{TC3}} & \multirow{2}{*}{\textbf{400}} & \textbf{Uncov. Flts.} & 115.50 $\pm$ 151.69 & 0 & 116 & 132 & 14 & 410 & 403 & 24 & 6 & 30 & 20 \\ 
 & & \textbf{Runtime (s)} & 6509.77 $\pm$ 2072.22 & 293.14 & 7200.28 & 7200.04 & 7200.61 & 7200.71 & 7200.53 & 7200.65 & 7200.4 & 7200.68 & 7200.67 \\ 
\cmidrule(l){2-14} 
& \multirow{2}{*}{\textbf{500}} & \textbf{Uncov. Flts.} & 0.0 $\pm$ 0.0 & 0 & \cellcolor{red!20}0 & 0 & 0 & 0 & \cellcolor{green!25}0 & 0 & 0 & 0 & 0 \\ 
 & & \textbf{Runtime (s)} & 333.93 $\pm$ 193.82 & 319.97 & \cellcolor{red!20}808.27 & 180.79 & 543.33 & 314.8 & \cellcolor{green!25}167.76 & 192.65 & 172.13 & 254.12 & 385.44 \\ 
\cmidrule(l){2-14} 
& \multirow{2}{*}{\textbf{600}} & \textbf{Uncov. Flts.} & 0.0 $\pm$ 0.0 & 0 & 0 & 0 & 0 & 0 & 0 & \cellcolor{green!25}0 & \cellcolor{red!20}0 & 0 & 0 \\ 
&  & \textbf{Runtime (s)} & 181.18 $\pm$ 60.07 & 133.42 & 169.46 & 155.78 & 122.96 & 233.31 & 139.05 & \cellcolor{green!25}122.13 & \cellcolor{red!20}306.18 & 258.16 & 171.34 \\ 
\cmidrule(l){2-14} 
& \multirow{2}{*}{\textbf{700}} & \textbf{Uncov. Flts.} & 0.0 $\pm$ 0.0 & 0 & 0 & 0 & 0 & \cellcolor{red!20}0 & 0 & 0 & 0 & 0 & \cellcolor{green!25}\textbf{0} \\ 
 & & \textbf{Runtime (s)} & 136.98 $\pm$ 57.98 & 235.08 & 102.95 & 122.92 & 111.77 & \cellcolor{red!20}264.42 & 119.76 & 94.14 & 110.94 & 123.44 & \cellcolor{green!25}\textbf{84.42} \\ 
\cmidrule(l){2-14} 
& \multirow{2}{*}{\textbf{800}} & \textbf{Uncov. Flts.} & 0.0 $\pm$ 0.0 & 0 & 0 & 0 & 0 & 0 & \cellcolor{green!25}\textbf{0} & 0 & \cellcolor{red!20}0 & 0 & 0 \\ 
 & & \textbf{Runtime (s)} & 149.84 $\pm$ 44.29 & 175.09 & 160.34 & 111.93 & 145.21 & 147.59 & \cellcolor{green!25}\textbf{92.34} & 107.01 & \cellcolor{red!20}260.2 & 155.67 & 142.98 \\ 
\cmidrule(l){2-14} 
& \multirow{2}{*}{\textbf{900}} & \textbf{Uncov. Flts.} & 0.0 $\pm$ 0.0 & 0 & \cellcolor{green!25}0 & 0 & 0 & 0 & 0 & 0 & 0 & \cellcolor{red!20}0 & 0 \\ 
 & & \textbf{Runtime (s)} & 293.21 $\pm$ 69.55 & 256.09 & \cellcolor{green!25}221.22 & 255.35 & 334.23 & 234.84 & 393.11 & 237.46 & 240.12 & \cellcolor{red!20}426.57 & 333.08 \\ 
\cmidrule(l){2-14} 
& \multirow{2}{*}{\textbf{1000}} & \textbf{Uncov. Flts.} & 0.0 $\pm$ 0.0 & 0 & 0 & 0 & 0 & 0 & 0 & \cellcolor{green!25}0 & 0 & \cellcolor{red!20}0 & 0 \\ 
 & & \textbf{Runtime (s)} & 656.37 $\pm$ 179.66 & 944.62 & 579.12 & 492.01 & 543.48 & 783.31 & 587.66 & \cellcolor{green!25}381.04 & 621.88 & \cellcolor{red!20}966.31 & 664.3 \\ 
\cmidrule(l){2-14} 
& \multirow{2}{*}{\textbf{1100}} & \textbf{Uncov. Flts.} & 0.0 $\pm$ 0.0 & 0 & \cellcolor{green!25}0 & 0 & \cellcolor{red!20}0 & 0 & 0 & 0 & 0 & 0 & 0 \\ 
 & & \textbf{Runtime (s)} & 2701.44 $\pm$ 849.14 & 1816.02 & \cellcolor{green!25}1219.25 & 3449.42 & \cellcolor{red!20}3930.72 & 3188.87 & 3401.02 & 2471.46 & 3432.76 & 2081.01 & 2023.83 \\ 
\cmidrule(l){2-14} 
& \multirow{2}{*}{\textbf{1200}} & \textbf{Uncov. Flts.} & 2.10 $\pm$ 1.76 & 1 & 3 & 3 & 6 & 4 & 1 & 1 & 1 & 0 & 1 \\ 
 & & \textbf{Runtime (s)} & 7733.26 $\pm$ 1848.75 & 7225.41 & 8212.13 & 9354.2 & 7438.45 & 8050.76 & 7559.46 & 7841.17 & 10370.6 & 2890.92 & 8389.52 \\
\bottomrule
\end{tabular}}
\vspace{-3.5mm}
\flushleft \scriptsize{For each setting of $K$ in which all random seeds lead to a feasible solution, the best and worst runtime values are highlighted in green and red colors respectively. Moreover, the best runtime values (approximately equal) for individual test cases are highlighted in bold.}
\label{ipdchResults}
\caption{Results of IFS generation for all test cases using the proposed IPDCH}
\end{center}
\end{table*}
\begin{itemize}
    \item The lower setting ($K=400$ flights) is chosen to ensure the presence of legal flight connections in the decomposed flight subsets, while preventing the number of IPDCH iterations from increasing drastically. This helps in controlling the runtime of the process from increasing drastically.
    \item The higher setting ($K=1200$ flights) is chosen such that the number of pairings being generated in each IPDCH-iteration does not increase drastically, keeping control over the IPDCH-iteration's runtime.
\end{itemize}
Moreover, IPDCH involves random selection of flights in each iteration, making it imperative to study the impact of pseudo-random number seed on IFS generation while finding the best-setting of $K$. For this, $10$ random seeds (varying uniformly) are used for each IPDCH-run corresponding to each setting of $K$. In this analysis, each IPDCH-run is terminated as soon as all given flights are covered, or its runtime exceeds $7200$ seconds (two hours) in case a feasible solution is not found. 
\par
\begin{figure}[pos=htb]
\begin{center}
\begin{subfigure}[b]{0.328\columnwidth}
    \centering
    \includegraphics[width=\linewidth, keepaspectratio]{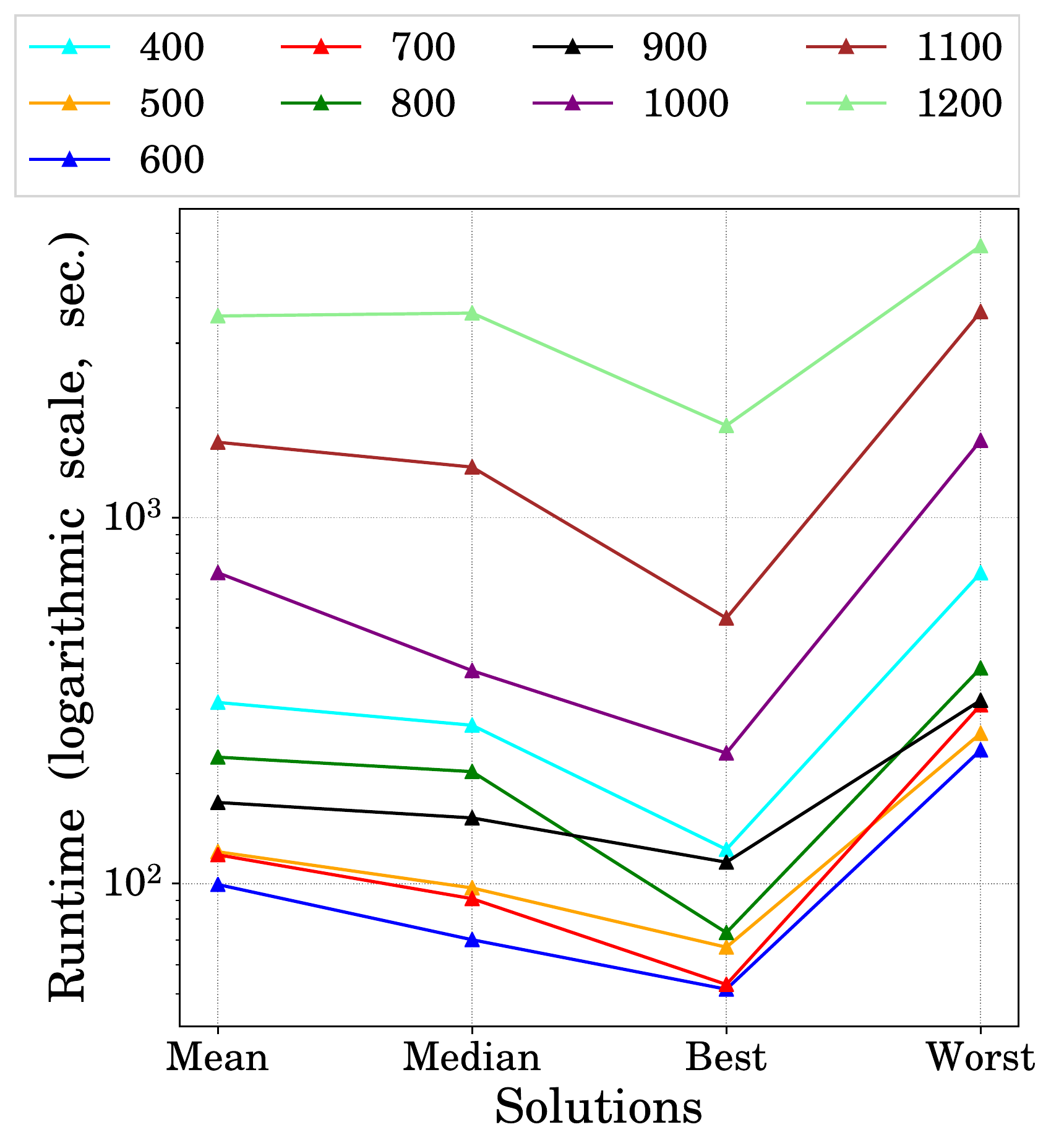}
    \caption{TC1}
    \label{graphlogscale1}
\end{subfigure}
\hfill
\begin{subfigure}[b]{0.317\columnwidth}
    \centering
    \includegraphics[width=\linewidth, keepaspectratio]{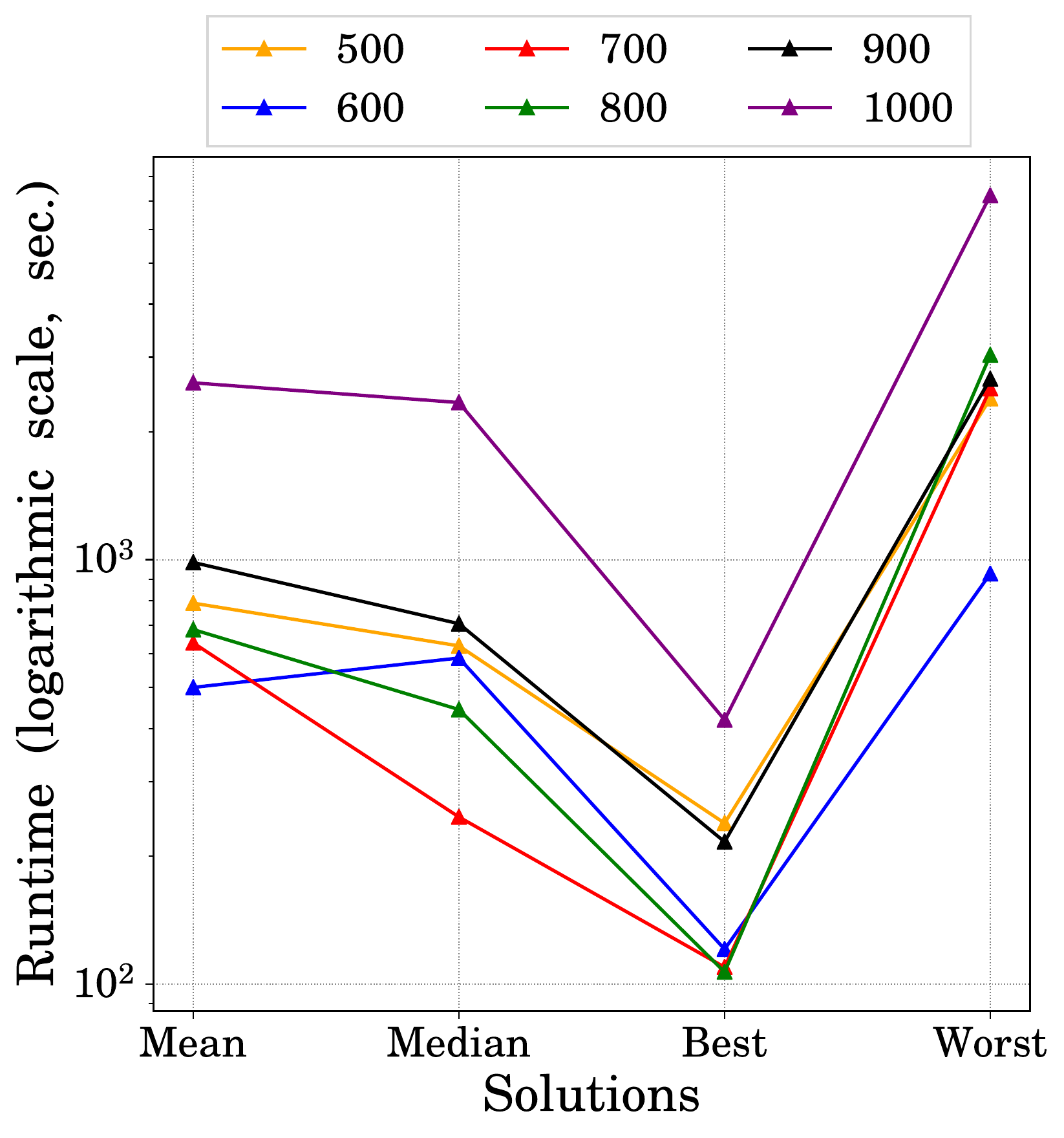}
    \caption{TC2}
    \label{graphlogscale2}
\end{subfigure}
\hfill
\begin{subfigure}[b]{0.328\columnwidth}
    \centering
    \includegraphics[width=\linewidth, keepaspectratio]{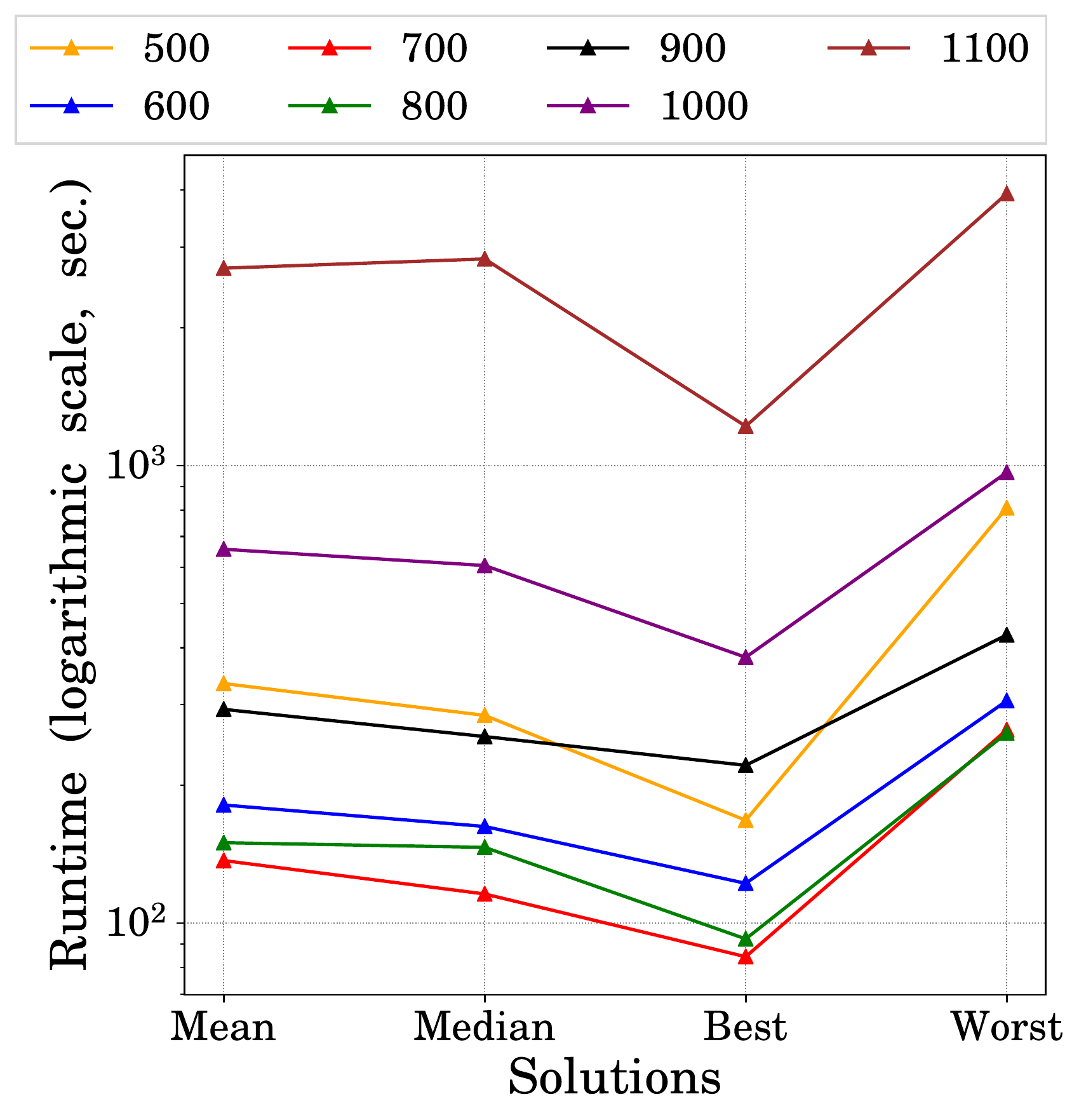}
    \caption{TC3}
    \label{graphlogscale3}
\end{subfigure}
\label{graphsAll}
\caption{Plots between mean, median, best, and worst runtime values for each setting of $K$ for each test case}
\end{center}
\end{figure}

 The IFS generation results (uncovered flights and runtime values) for all test cases using the proposed IPDCH with the above mentioned experimental settings are presented in Table~\ref{ipdchResults}. In the table, for each setting of $K$, the results corresponding to each random seed are presented in each column along with a column with ($\overline{x} \pm \sigma$) values. Some of the observations drawn from this table are now discussed. For TC1, it is noted that IPDCH-runs with all values of $K$ lead to a feasible solution for all random seeds. However, for TC2, IPDCH-runs with $K = 400, 1000, 1100$ \& $1200$ flights are not able to generate a feasible solution for some of their random seeds in the given runtime. Similarly, for TC3, IPDCH-runs with $K = 400$ \& $1200$ flights are not able to generate a feasible solution for some of their random seeds in the given runtime. Hence, the best setting of $K$ lies in the range of $500-900$ flights. For further visualization, a comparative plot is drawn in between mean, median, best, and worst runtime values for each feasible setting of $K$ for each test case, as shown in figure~\ref{graphsAll}.
From these plots and the results table, it is clear that the best IPDCH-run for TC1 is for $K = 600$ \& $700$ flights. Similarly, for TC2 \& TC3, the same is observed for $K= 700$ \& $800$ flights. For each test case, two best settings of $K$ are mentioned as the difference between their best-runtime values is marginal. It is to be noted that all these results are dependent on the computational resources used in this work (Section~\ref{sec:settings}). Hence, for the given configuration of computational resources and the characteristics of the flight data sets used, it could be concluded that $K=700$ flights is the best-setting of the proposed IPDCH for initializing large-scale and complex CPOPs in a cost-effective and time-efficient manner. 
In the following subsection, the performance of the proposed IPDCH is compared with an existing IFS generation method from the literature.
\subsubsection{Comparison between IPDCH \& Enhanced-DFS Heuristic} \label{enhancedDFSsec}
To demonstrate the utility of the proposed IPDCH, its performance is compared with the Enhanced-DFS heuristic (authors' previous IFS-generation heuristic) and the results are presented. For this comparison, the setting of $K$ for IPDCH-runs is kept at 700 flights as concluded in the previous subsection. The IFS generation results for both of these heuristics are summarized in Table~\ref{enhancedDFS}.
\begin{table*}[pos=htbp]
\scriptsize
\begin{center}
\begin{tabular}{M{5mm}M{23.5mm}M{13mm}M{8mm}M{10mm}}
\toprule
\textbf{Test} & \textbf{IFS Generation} & \textbf{\# Pairings} &\textbf{Runtime}& \textbf{LP-cost}\\
\textbf{Case} & \textbf{Method} & &\textbf{(sec.)}& \textbf{(USD)}\\
\toprule
\multirow{2}{*}{\textbf{TC1}} & IPDCH with $K=$ 700 & 4,410 & \cellcolor{green!25}53 & 3,743,402\\
\cmidrule(l){2-5}
& Enhanced-DFS & 285,057 & 1771 & 2,824,341\\
\midrule
\multirow{2}{*}{\textbf{TC2}} & IPDCH with $K=$ 700 & 7,991 & \cellcolor{green!25}110 & 4,696,619\\
\cmidrule(l){2-5}
& Enhanced-DFS & 477,617 & 6625 & 3,646,715\\
\midrule
\multirow{2}{*}{\textbf{TC3}} & IPDCH with $K=$ 700 & 6,056 & \cellcolor{green!25}85 & 5,002,751\\
\cmidrule(l){2-5}
& Enhanced-DFS & 26,678 & 108 & 4,971,053\\
\bottomrule
\end{tabular}
\label{enhancedDFS}
\caption{Comparison of IFSs generated using the proposed IPDCH and the Enhanced-DFS heuristic}
\end{center}
\end{table*}
It is observed that the propsoed IPDCH took only up to 110 seconds to generate an IFS for the most complex and/or largest test cases used in this work. On the contrary, the respective runtime values for Enhanced-DFS heuristic are up to $\sim$6625 seconds ($=$1.85 hours). The difference between their runtime performance is marginal for TC3 but is significantly huge for TC1 \& TC2. These observations validate that the proposed IPDCH (with $K = 700$ flights) is a fast initialization heuristic in comparison to the Enhanced-DFS. TC1 \& TC2 are not only derived from complex flight networks but also contain some flights with rare legal flight connections and are extremely difficult to be identified by an initialization-search. This is the reason behind the huge runtime values of the Enhanced-DFS heuristic. However, the performance of the proposed IPDCH remains unaffected even for such flight networks, making it one of the most time-efficient IFS generation heuristic present for large-scale CPOPs. In addition to this, the cost-quality of an IFS generated using IPDCH could be improved by running the IPDCH for a longer time. Hence, for an IPDCH-run, it is imperative to analyze the trade-off between the cost improvement it brings in each iteration and its overall runtime which is presented in the following subsection. Given this trade-off, it becomes critically important to time the termination of an IPDCH-run as the IFS characteristics (cost and DOSF) are directly dependent on these termination choices. Furthermore, it is anticipated that the success of the subsequent optimization phase is linked with the interplay of these two characteristics. For this, the impact of IFS characteristics on final crew pairing solution is discussed in the following subsection.
\subsection{Sensitivity of IFS charanteristics on the performance of $\bm{AirCROP}$} \label{sensitivity}
In Section~\ref{methods}, two important characteristics of an IFS (cost and DOSF) are discussed. It is anticipated that the success of subsequent crew pairing optimization phase is linked to the interplay between these characteristics. To establish this empirically, a sensitivity analysis of IFS characteristics on final crew pairing solutions, obtained using $AirCROP$, is presented in this subsection. For this, multiple IFSs (with varying characteristics) are generated using the proposed IPDCH. Subsequently these IFSs are individually fed as input to the developed optimizer, and the characteristics of the final crew pairing solutions (their cost and runtime values) are analyzed.
\par
\begin{figure}[pos=htbp]
\centering{\includegraphics[width=0.45\columnwidth, keepaspectratio]{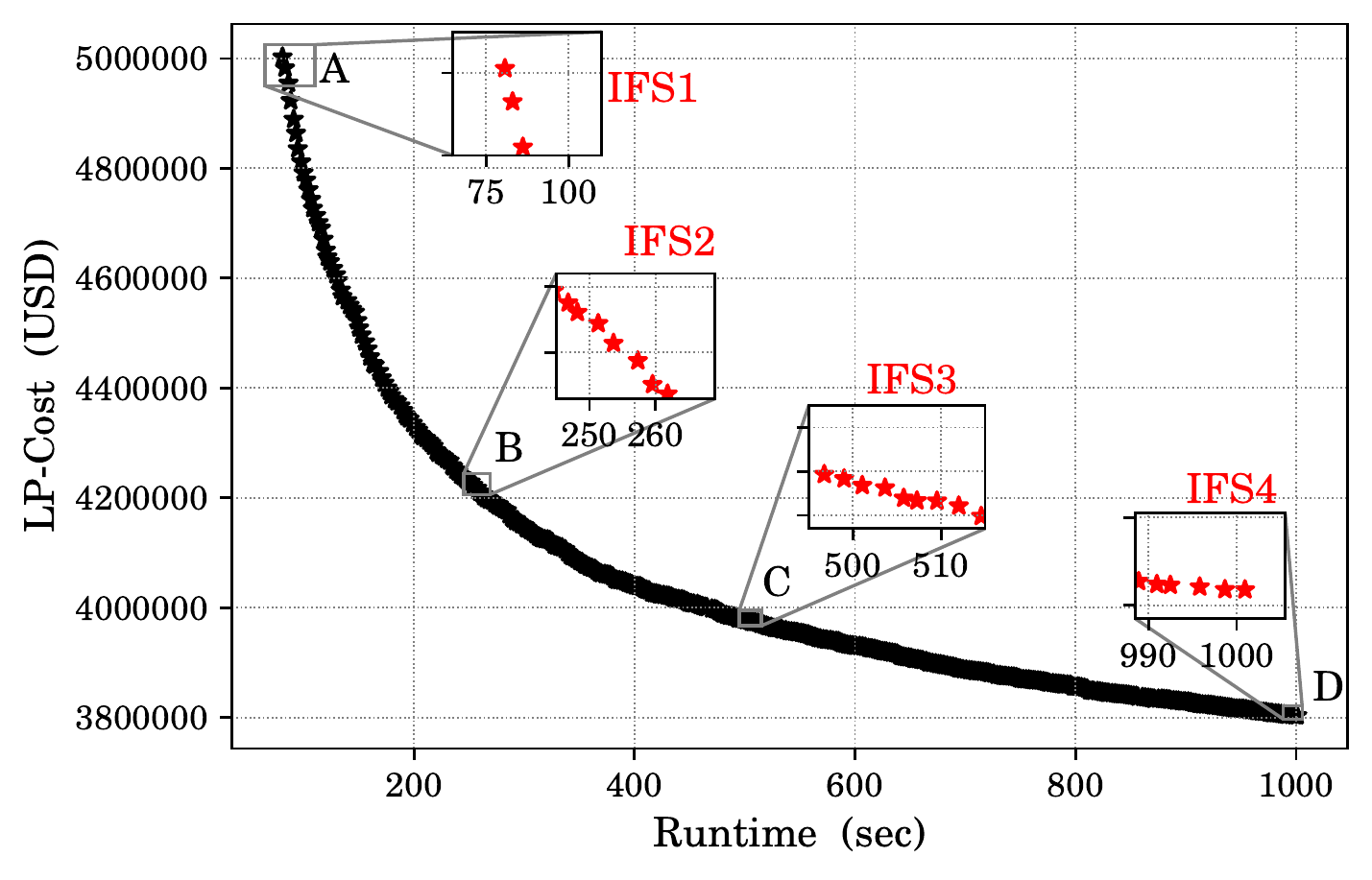}
\caption{Plot of variation in LP-cost of the IFS with runtime of an IPDCH-run for TC3}}
\label{costvsruntimetradeoff}
\end{figure}

 In Section~\ref{methods}, it is also discussed that cost and DOSF have complimentary relationships with the runtime of an IPDCH-run, and could be regulated by varying its termination point. To study the trade-off between cost-improvement with each IPDCH-iteration and its total runtime, a long IPDCH-run is performed for TC3 with $K =700$ flights and runtime $=1000$ seconds as the termination criterion. Using data of this run, the LP-cost of the IFS after each IPDCH-iteration is plotted against the respective runtime values (measured from the beginning of the run) which is shown in fig.~\ref{costvsruntimetradeoff}.
It is to be noted that in this figure, the first data point belongs to an IPDCH-iteration immediately after feasibility point is achieved, i.e., after 85 seconds for TC3. From this graph, it is evident that the cost of an IFS is inversely proportional to the IPDCH-runtime, i.e., the higher the runtime of an IPDCH-run, the lower would be the LP-cost of the resulting IFS and vice versa. As shown in the figure, an IFS with best cost-quality is selected from region \textit{D}, i.e., after the iteration with runtime $ =1000$ seconds. And, an IFS with worst cost-quality is selected from the region \textit{A}, i.e., after an iteration with runtime $= 85$ seconds. The DOSF of an IFS also varies inversely with the IPDCH-runtime. The IFS from region \textit{A} will have highest DOSF whereas the IFS from region \textit{D} will have the lowest DOSF. As mentioned above, multiple IFSs with varying characteristics are required to carry out the sensitivity analysis of an IFS characteristics on the final crew pairing solution. For this, two more IFSs (after the ones selected from the regions \textit{A} \& \textit{D}) are selected from two other complimentary regions of this figure, i.e., regions \textit{B} \& \textit{D}. The rationale behind selecting these other two regions is as follows. The trade-off in this figure reveals that the cost improvement per iteration is huge in initial iterations of the IPDCH-run (from regions \textit{A} to \textit{C}) and it decreases with further iterations (from regions \textit{C} \& \textit{D}). In addition to these three regions, the fourth region (\textit{B}) is selected as the mid-iterations between regions \textit{A} \& \textit{C}. In a similar manner, multiple IFSs are generated for TC1 \& TC2  which are summarized in Table~\ref{finalSolutionsTable}.
\begin{table*}[pos=htbp]
\scriptsize
\begin{center}
\begin{tabular}{M{4mm}M{5mm}M{10mm}M{8mm}M{10mm}M{19mm}M{13.5mm}M{10mm}M{0.2mm}M{13.5mm}M{10mm}}
\toprule
\multirow{1.66}{*}{\textbf{Test}} & \multirow{5}{*}{\textbf{IFS(s)}} & \multirow{5}{*}{\textbf{\# Pairings}} & \multirow{3.5}{*}{\textbf{Runtime}} & \multirow{3.5}{*}{\textbf{LP-cost}} & \multirow{5}{*}{\textbf{DOSF}}  &  \multicolumn{5}{c}{\textbf{Final Crew Pairing Solution}} \\
\cmidrule{7-11}
\multirow{1.67}{*}{\textbf{Case}} & & & \multirow{3.5}{*}{\textbf{(sec.)}} & \multirow{3.5}{*}{\textbf{(USD)}} & & \multicolumn{2}{c}{\textbf{I Run}} & & \multicolumn{2}{c}{\textbf{II Run}} \\
\cmidrule{7-8} \cmidrule{10-11}
\multirow{1.66}{*}{\textbf{(TC)}} & & & & & & \textbf{Total Cost} & \textbf{Runtime} & & \textbf{Total Cost} & \textbf{Runtime} \\
& & & & & & \textbf{(USD)} & \textbf{(hours)} & & \textbf{(USD)} & \textbf{(hours)} \\ 
\midrule
\multirow{4}{*}{\textbf{TC1}} & $\mathcal{IFS}1$ & 4,410 & 53 & \cellcolor{blue!5}3,743,402 & \cellcolor{green!45}High & 2678720 & 2.8425 &  & 2684207 & 3.2453 \\
                     & $\mathcal{IFS}2$ & 18,272  & 250 & \cellcolor{blue!15}3,064,397 & \cellcolor{green!30}Moderately High & 2696991 & 3.3542 &  & 2684624 & 3.7778 \\
                     & $\mathcal{IFS}3$ & 36,965 & 500 & \cellcolor{blue!30}2,915,924 & \cellcolor{green!15}Moderately Low & 2693286 & 3.4487 &  & 2683362 & 3.8042 \\
                     & $\mathcal{IFS}4$ & 79,397 & 1000 & \cellcolor{blue!45}2,819,007 & \cellcolor{green!5} Low & 2698492 & 3.4362 &  & 2694067 & 3.9000\\
\midrule
\multirow{4}{*}{\textbf{TC2}} & $\mathcal{IFS}1$ & 7,991 & 110 & \cellcolor{blue!5}4,696,619 & \cellcolor{green!45}High & 3511944 & 6.0362 &  & 3502514 & 6.5278 \\
                     & $\mathcal{IFS}2$ & 18,686 & 250 & \cellcolor{blue!15}4,158,096 & \cellcolor{green!30}Moderately High & 3497212 & 5.9598 &  & 3487858 & 6.3473 \\
                     & $\mathcal{IFS}3$ & 36,458 & 500 & \cellcolor{blue!30}3,932,366 & \cellcolor{green!15}Moderately Low & 3494837 & 6.2848 &  & 3492258 & 6.6153 \\
                     & $\mathcal{IFS}4$ & 70,285 & 1000 & \cellcolor{blue!45}3,775,848 & \cellcolor{green!5} Low & 3489654 & 6.2848 &  & 3481849 & 6.6042\\
\midrule
\multirow{4}{*}{\textbf{TC3}} & $\mathcal{IFS}1$ & 6,056 & 85 & \cellcolor{blue!5}5,002,751 & \cellcolor{green!45}High & 3513784 & 5.5500 &  & 3513437 & 5.8167 \\
                     & $\mathcal{IFS}2$ & 17,747 & 250 & \cellcolor{blue!15}4,238,840 & \cellcolor{green!30}Moderately High & 3534496 & 6.3028 &  & 3514034 & 6.7278 \\
                     & $\mathcal{IFS}3$ & 36,306 & 500 & \cellcolor{blue!30}3,979,340 & \cellcolor{green!15}Moderately Low & 3517987 & 5.7625 &  & 3512787 & 6.2625 \\
                     & $\mathcal{IFS}4$ & 73,827 & 1000 & \cellcolor{blue!45}3,803,569 & \cellcolor{green!5} Low & 3534042 & 6.3848 &  & 3538691 & 6.7459\\
\bottomrule
\end{tabular}
\label{finalSolutionsTable}
\caption{Results of sensitivity analysis of an IFS characteristics on the final crew pairing solution}
\end{center}
\end{table*}
\par Now, these IFSs are fed as input to the developed optimizer for finding the respective final crew pairing solutions. In optimization systems, trade-off between \textit{solution-quality} (here, cost) and \textit{runtime} plays a critical role in timing its termination. Generally, in small-scale optimization systems, solution quality is preferred over runtime as their search-efficiency largely remains unaffected by the problem scale. This is opposite for large-scale optimization systems as their search efficiency is marred by the curse of dimensionality (vast search space). The $AirCROP$ belongs to the category of large-scale optimization systems, hence, its choice of termination lies with airline users. However, to accommodate the varying operational-scales of the airline users, the developed optimizer is designed with multiple termination points. For this research, two different termination points of the developed optimizer are used: \textit{Setting-1} (to find an approximately good-cost final crew pairing solution in less runtime) \& \textit{Setting-2} (to find better-cost final crew pairing solution by spending more runtime). For simplicity, the optimizer run with Setting-1 is referred to as \textit{I Run}, and the optimizer run with Setting-2 is referred to as \textit{II Run}.
\par
\begin{figure}[pos=htbp]
\centering
\begin{subfigure}[b]{0.325\columnwidth}
    \centering
    \includegraphics[width=\linewidth, keepaspectratio]{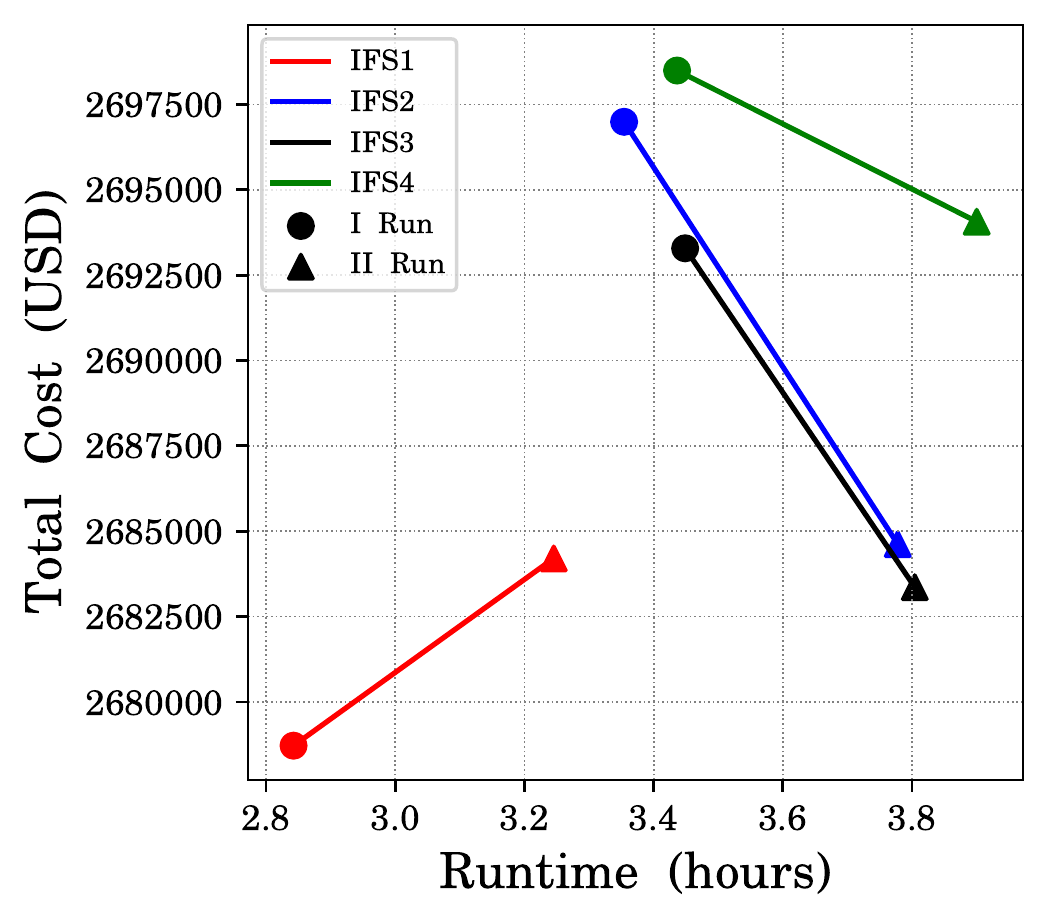}
    \caption{TC1}
    \label{finalSolutionsGraph1}
\end{subfigure}
\hfill
\begin{subfigure}[b]{0.325\columnwidth}
    \centering
    \includegraphics[width=\linewidth, keepaspectratio]{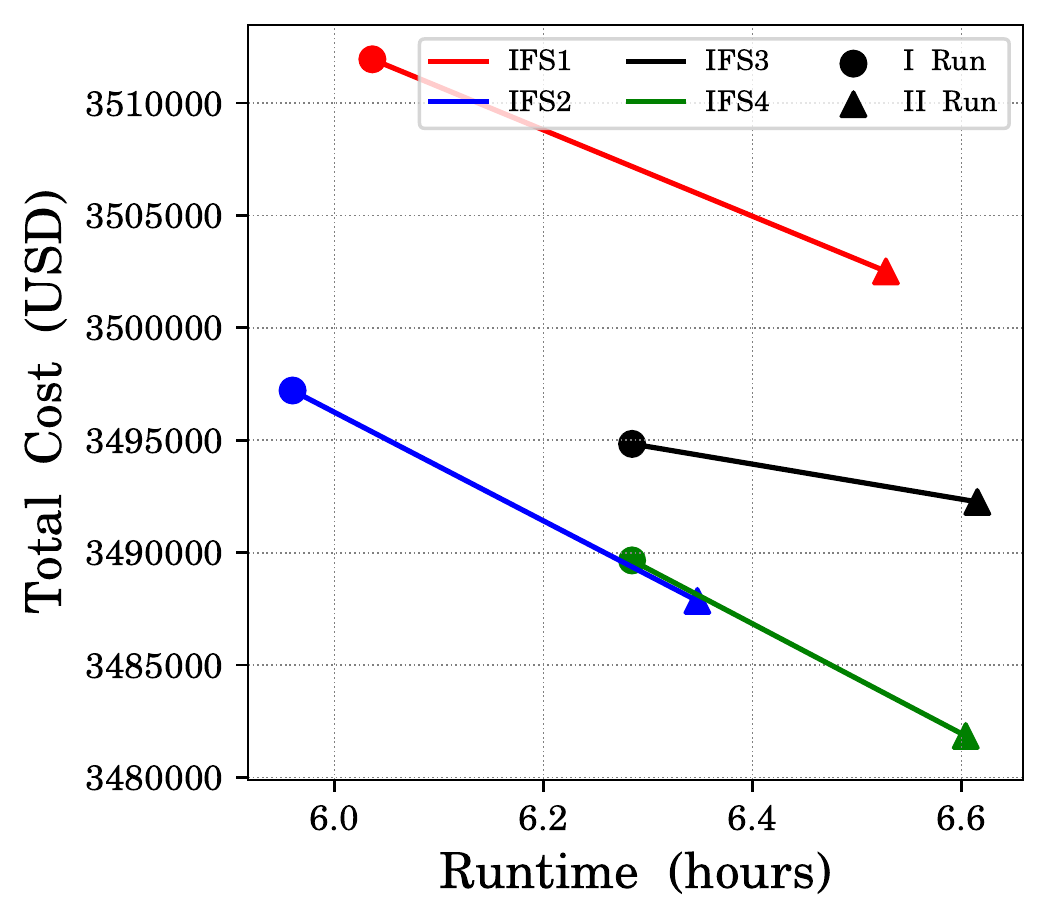}
    \caption{TC2}
    \label{finalSolutionsGraph2}
\end{subfigure}
\hfill
\begin{subfigure}[b]{0.325\columnwidth}
    \centering
    \includegraphics[width=\linewidth, keepaspectratio]{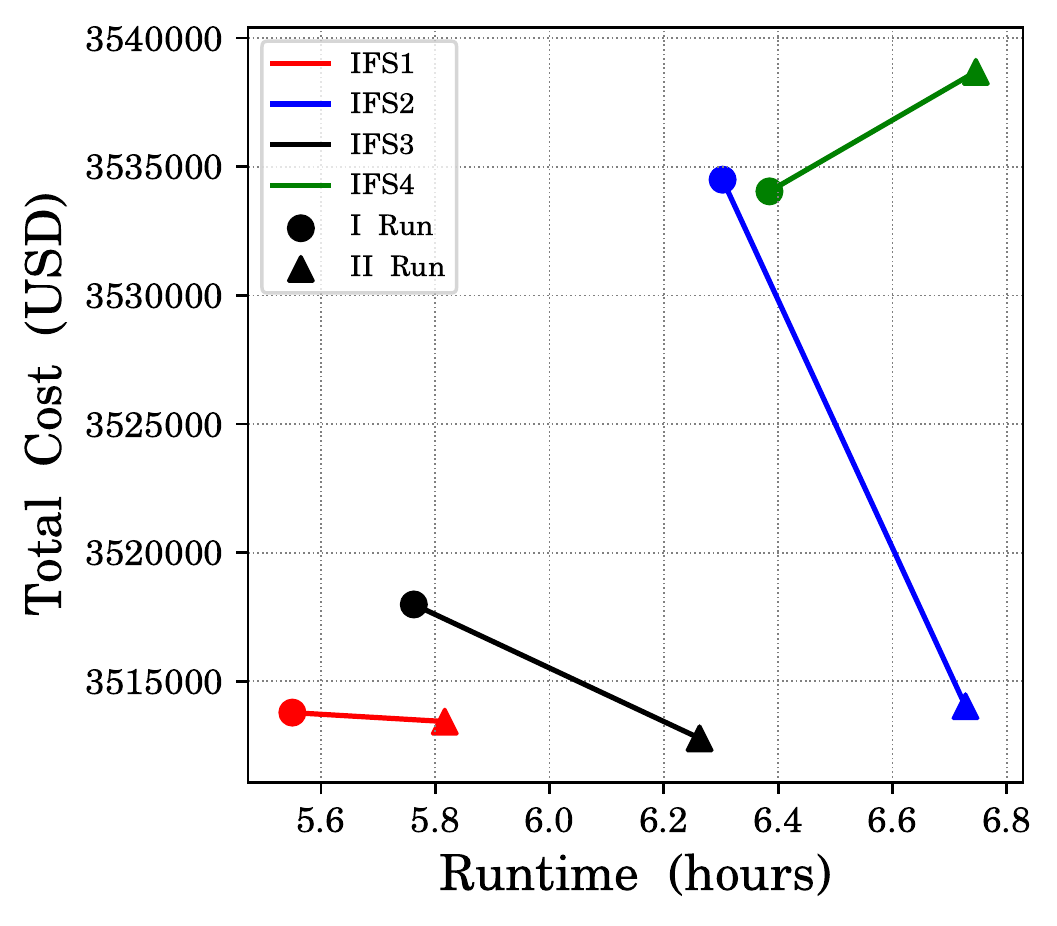}
    \caption{TC3}
    \label{finalSolutionsGraph3}
\end{subfigure}
\caption{Plots between cost and runtime values of final crew pairing solutions generated from IFSs with varying characteristics}
\label{graphsAll2}
\end{figure}

 For all test cases, results of optimizer runs with each of the above-generated IFSs as input are summarized in Table~\ref{finalSolutionsTable}. The quality of a final crew pairing solution is measured in terms of its final cost and the runtime spent in finding that solution (including IPDCH-runtime). These results are visualized by plotting cost against runtime values for each test case, as shown in figures~\ref{finalSolutionsGraph1}, ~\ref{finalSolutionsGraph2} \& ~\ref{finalSolutionsGraph3}.
Before summarizing the observations from these results, it is imperative to introduce \textit{non-dominated solutions}. \citet{deb2001multi} defined a domination criterion which states that a solution $x_1$ dominates another solution $x_2$ only if $f_i(x_1) \leq f_i(x_2)$ for all functions $f_i$, and $f_i(x_1) < f_i(x_2)$ for at least one $f_i$. And, the non-dominated solutions are those solutions which are not dominated by another solution.
The observations drawn from the results of this sensitivity analysis are as follows:
\begin{itemize}
\item As mentioned-above, I Run is linked to the generation of an approximately good-cost solution in \textit{less} runtime whereas, the II Run is linked to the generation of a \textit{better-cost} solution by spending more runtime. The ranking of final crew pairing solutions for I Run (i.e. w.r.t. the runtime) is as follows:
\begin{equation}
\begin{aligned}
\small For\ TC1&: R_{\mathcal{IFS}1} > R_{\mathcal{IFS}2} > R_{\mathcal{IFS}3} > R_{\mathcal{IFS}4}\\
\small For\ TC2&: R_{\mathcal{IFS}2} > R_{\mathcal{IFS}1} > R_{\mathcal{IFS}4} > R_{\mathcal{IFS}3}\\
\small For\ TC3&: R_{\mathcal{IFS}1} > R_{\mathcal{IFS}3} > R_{\mathcal{IFS}2} > R_{\mathcal{IFS}4}
\end{aligned}
\end{equation}
and the same for II Run (i.e. w.r.t. the cost) is:
\begin{equation}
\begin{aligned}
\small For\ TC1: R_{\mathcal{IFS}3} > R_{\mathcal{IFS}1} > R_{\mathcal{IFS}2} > R_{\mathcal{IFS}4}\\
\small For\ TC2: R_{\mathcal{IFS}4} > R_{\mathcal{IFS}2} > R_{\mathcal{IFS}3} > R_{\mathcal{IFS}1}\\
\small For\ TC3: R_{\mathcal{IFS}3} > R_{\mathcal{IFS}1} > R_{\mathcal{IFS}2} > R_{\mathcal{IFS}4}
\end{aligned}
\end{equation}
\item For TC1 \& TC3, the best solution of I Run is obtained using $\mathcal{IFS}1$ as input. This solution is not only best in terms of overall runtime but is also best in terms of cost. However, for their respective II-Runs, non-dominated solutions are obtained using $\mathcal{IFS}1$ \& $\mathcal{IFS}3$ as input. Among these non-dominated solutions, it is found that the cost of $\mathcal{IFS}1$'s final solution is marginally poorer than that of the $\mathcal{IFS}3$ (0.032\% for TC1 and 0.019\% for TC3). However, the difference between runtime values of the former and latter solutions is huge (17.22\% for TC1 and 7.67\% for TC3). Similarly, for TC2, two non-dominated solutions are obtained in both optimizer runs using $\mathcal{IFS}2$ \& $\mathcal{IFS}4$ as input. The cost of $\mathcal{IFS}2$'s final solution is marginally lower than that of the $\mathcal{IFS}4$ (0.22\% in I Run and 0.18\% in II Run) whereas, the difference between their respective runtime values is huge (5.5\% in I Run and 4.1\% in II Run). From these observations, it is clear that for initializing comparable large-scale CPOPs, $\mathcal{IFS}1$/$\mathcal{IFS}2$ is a better choice. These IFSs are generated using IPDCH with $K = 700$ flights \& up to $250$ seconds of runtime, and in turn have high/moderately high DOSF. Hence, it would be appropriate to conclude that an IFS with high/moderately high DOSF drives the optimization search towards near-optimal solutions for large-scale CPOPs in a time-efficient manner.
\item In the majority of these results, it is also observed that if I Run did not lead to a good-cost solution, then the II Run results in huge cost reductions. However, this cost reduction becomes marginal if a good-cost solution is already obtained in the I Run. Keeping in mind the user preferences for large-scale optimization systems, it is desirable to select an IFS which drives the optimization search to a good-cost solution in \textit{less} runtime, i.e., the result of I Run. These IFSs are either $\mathcal{IFS}1$ (in case of TC2 \& TC3) or $\mathcal{IFS}2$ (in case of TC2). Hence, this provides further evidence in support of the conclusion drawn in the previous point.
\item For TC1 with $\mathcal{IFS}1$ as input, the cost-reduction in the final solution of II Run w.r.t. the I Run becomes negative. A similar pattern was observed with $\mathcal{IFS}4$ as input for TC3. This reveals the complicated nature of the problem at hand. In these kind of problems, where an integer solution is obtained by first optimizing the problem in \textit{continuous} domain followed by optimization in \textit{integer} domain, it is not guaranteed that the integer solution will always improve with the runtime, making it imperative to time the termination decisions.
\end{itemize}

\section{Conclusion and Future Research} \label{conc}
For large-scale and complex airline CPOPs, IFS generation standalone is computationally challenging as it constitutes an NP-complete problem. In this research, a cost-effective and time-efficient initialization heuristic, IPDCH, is proposed. Its utility is demonstrated on real-world, large-scale (over 3228 flights), complex airline test cases (with 15 crew bases), provided by GE Aviation. The proposed IPDCH not only utilizes a divide-and-cover strategy (the input flight schedule is divided into smaller flight subsets) to achieve feasibility in a time-efficient manner, but also attempts to cover these flight subsets in a cost-effective manner by optimization using IP. For this, the IPDCH employs the decomposition parameter $K$ which controls the size of the flight subsets. To leverage-in the maximum gain in search-efficiency of the IPDCH, it is imperative to tune the setting of $K$ for the given computational resources and airline flight networks. As a conclusion of this research, it is advisable for airline users to keep $K=700$ flights in IPDCH for initializing their comparable flight networks using similar computational resources.
\par In this paper, the proposed IPDCH is compared with an existing IFS generation heuristic from the literature (Enhanced-DFS heuristic). It was observed that the IPDCH took up to only 110 seconds for generating an IFS for the most complex and/or largest test case used whereas, the Enhanced-DFS heuristic took up to 6625 seconds (1.85 hours) for the same test cases respectively. This shows that for the most complex test cases involved, the Enhanced-DFS heuristic becomes an exhaustive search, revealing its dependency on the characteristics of the flight data set. However, even for such test cases, the proposed IPDCH generates an IFS in few seconds. This establishes the superiority of the proposed IPDCH over the best IFS generation heuristic available in the literature for initializing large-scale and complex CPOPs.
\par An IFS is used to initialize the crew pairing optimization phase, and intuitively, it is desirable to have an IFS which is flexible enough to promote an exploratory initial optimization search instead of the exploitative. To establish this empirically, the sensitivity of the IFS characteristics (identified as cost and DOSF) on the final crew pairing solution is analyzed in this paper. Surprisingly, the literature is silent on such empirical evidence which is otherwise critically important for the success of an initialization heuristic to result in a good-cost final crew pairing solution in less runtime. From this sensitivity analysis, it is concluded that an IFS with high/moderately high DOSF drives the optimization towards the near-optimal final crew pairing solution in less runtime. For similar computational resources as used in this research, these IFSs could be obtained using IPDCH with $K=700$ flights and runtime up to $250$ seconds.
However, the best choice for an airline user is to terminate the IPDCH at the feasibility point in order to avoid the empirical study with their own computational resources and data sets.
\par As future work, a plausible research direction for the proposed IPDCH could be to establish the relationship of $K$ with the configuration of computational resources being used and with the characteristics of flight data sets being solved. Another research direction would be to replace the random decomposition strategy with a knowledge heuristic that could use the information of legal flight connections while decomposing the input flight schedule. This may increase the probability of covering more unique flights in lesser IPDCH-iterations. Furthermore, for a given airline, the final crew pairing solutions from hundreds of previous optimizer runs could be utilize to learn flight-connection structures/patterns which could be used to warm-start the IFS generation. This will not only help in speeding up the IFS generation process but could also help in initializing the subsequent optimization phase from a critically important point in the search space.
%
%
\section*{Acknowledgment}
This work is a research outcome of an Indo-Dutch joint research project. It is supported by the Ministry of Electronics and Information Technology (MEITY), India [grant 13(4)/2015-CC\&BT]; Netherlands Organization for Scientific Research (NWO), the Netherlands; and General Electric (GE) Aviation, India. The authors would like to acknowledge the invaluable support of GE Aviation team members: Saaju Paulose (Senior Manager), Arioli Arumugam (Senior Director- Data \& Analytics), and Alla Rajesh (Senior Staff Data \& Analytics Scientist) for providing problem definition, real-world test cases, and for sharing domain-knowledge during numerous stimulating discussions which helped the authors in successfully completing this work.
%
%
\bibliographystyle{elsarticle-num-names-alphsort}
\bibliography{cas-refs}
\end{document}